%% file: paper.tex
\documentclass[review]{elsarticle}

\usepackage{lineno,hyperref}
\modulolinenumbers[5]

\journal{Elsevier}

\input{packages}
\input{commands}

\begin{document}

\begin{frontmatter}

\title{MPTP: Motion-Planning-aware Task Planning for Navigation in Belief Space}

\author{Antony Thomas, Fulvio Mastrogiovanni, Marco Baglietto}
\address{Department of Informatics, Bioengineering, Robotics, and Systems Engineering, University of Genoa, Via All'Opera Pia 13, 16145 Genoa, Italy. \\\  \textsl{\{antony.thomas@dibris.unige.it, fulvio.mastrogiovanni@unige.it, marco.baglietto@unige.it\}}}




\begin{abstract}
We present an integrated Task-Motion Planning (TMP) framework for navigation in large-scale  environments. Of late, TMP for manipulation has attracted significant interest resulting in a proliferation of different approaches. In contrast, TMP for navigation has received considerably less attention. Autonomous robots operating in real-world complex scenarios require planning in the discrete (task) space and the continuous (motion) space. In knowledge-intensive domains, on the one hand, a robot has to reason at the highest-level, \revtext{for example, the objects to procure, the regions to navigate to in order to acquire them}; on the other hand, the feasibility of the respective navigation tasks have to be checked at the execution level. This presents a need for motion-planning-aware task planners. In this paper, we discuss a probabilistically complete approach that leverages this task-motion interaction for navigating in large knowledge-intensive domains, returning a plan that is optimal at the task-level. The framework is intended for motion planning under motion and sensing uncertainty, which is formally known as belief space planning. The underlying methodology is validated in simulation, in an office environment \revtext{and its scalability is tested in the larger Willow Garage world. A reasonable comparison with a work that is closest to our approach is also provided}. We also demonstrate the adaptability of our approach by considering a building floor navigation domain. Finally, we also discuss the limitations of our approach and put forward suggestions for improvements and future work.
\end{abstract}

\begin{keyword}
Task-Motion Planning, Belief Space Planning, Autonomous Navigation
\end{keyword}

\end{frontmatter}


\section{Introduction}
\input{intro}

\label{sec:one}
\section{Related Work}
\input{related_work}

\section{Preliminaries and Definitions}
\label{sec:preliminaries}
\input{preliminaries}

\section{Approach}
\label{sec:approach}
\input{methodology}

\section{Implementation and Experimental Results}
\label{sec:results}
\input{results}

\section{Discussion}
\input{discussion}

\section{Conclusions}
\input{conclusion}

\section*{Acknowledgment}
We thank Chiara Piacentini for her valuable inputs on the POPF-TIF planner that were very helpful in our implementation. 
\section*{References}
\bibliographystyle{elsarticle-num}
\bibliography{/home/antony/Research_Genoa/References/References}

\end{document}

%% file: packages.tex
\usepackage{url}

\usepackage{type1cm}        
\usepackage{makeidx}         
\usepackage{graphicx}        
\usepackage{multicol}        
\usepackage[bottom]{footmisc}

\usepackage{newtxtext}       %
\usepackage{newtxmath}       
\usepackage{color,xcolor,ucs}
\usepackage{subfig}
\usepackage[font=small,labelfont=bf]{caption}
\usepackage{floatrow}
\usepackage{tabularx}
\usepackage{float}
\usepackage{amsmath}

\usepackage{xr-hyper}
\usepackage{wrapfig}

\usepackage{algorithm}
\usepackage{algpseudocode}
\usepackage{multirow}
\usepackage{listings}
\usepackage{lipsum}
\usepackage{textcomp} 

%% file: commands.tex
\makeatletter
\algnewcommand{\LineComment}[1]{\Statex \hskip\ALG@thistlm \(\triangleright\) #1}
\algnewcommand{\LineCommentCont}[1]{\Statex \hskip\ALG@thistlm \parbox[t]{\linegoal}{\hangindent=1em\hangafter=1 $\triangleright$ #1}}
\makeatother

\newdefinition{defn}{Definition}

\definecolor{mygray}{gray}{0.6}

\definecolor{revisioncolor}{rgb}{0,0,0}
\newcommand{\revtext}[1]{{\color{revisioncolor} #1}}

%% file: intro.tex
Autonomous robots operating in complex real world scenarios require different levels of planning to execute \revtext{the} assigned tasks. High-level (task) planning helps break down a given set of tasks into a sequence of sub-tasks. Actual execution of each of these sub-tasks would require low-level control actions to generate appropriate robot motions. In fact, the dependency between logical and geometrical aspects is pervasive in both task planning and execution. Hence, planning should be performed in the task-motion or the discrete-continuous space~\cite{lagriffoul2018RAL}. 

In recent years, combining high-level task planning with low-level motion planning has been a subject of great interest among the Robotics and Artificial Intelligence (AI) communities. Traditionally, task planning and motion planning have evolved as two independent fields. AI planning frameworks \revtext{such} as the Planning Domain Definition Language (PDDL)~\cite{mcdermott1998AIPS} mainly focus on high-level task planning supposing that the geometric preconditions (e.g., grasping poses for a pick-up task~\cite{srivastava2014ICRA}) for the robot motion to carry out these tasks are achievable. In reality, such an assumption can be catastrophic as an action or sequence of actions generated by the task planner might turn out to be unfeasible at the controller execution level. 

Over the past few years, Task-Motion Planning (TMP) for manipulation has received considerable interest among the research community~\revtext{\cite{srivastava2014ICRA, garrett2018IJRR, kaelbling2013IJRR, dantam2018IJRR,garrett2019arxiv}}. Robot-based manipulation domain calls for discrete and continuous reasoning to execute the required action reliably. For example, a simple table top domain requires the robot to reason at the discrete level to decide the objects to be picked up and also the order of these high-level actions. The execution of these discrete actions require continuous reasoning in the configuration space of the robot to generate appropriate motions. Yet, a discrete action might turn out to be unfeasible due to the end-effector's reachability workspace. This might be due to the availability of a partial map leading to unmodeled objects or occlusions leading to unobserved objects or simply because the robot is too close the target object, rendering a grasp action impossible. This presents the need for a tight coupling between task planning and motion planning, enabling an interface for efficient interaction between the symbolic and geometric layers. TMP for navigation presents different challenges in comparison to TMP for manipulation. As such, TMP for navigation has not yet received much attention and therefore lacks sufficient literature. TMP for navigation essentially involves \revtext{reasoning about different objects and their properties, deciding which objects to procure, selecting high-level actions that satisfy the low-level continuous motion constraints to navigate to the objects or other locations of interest, and finally procuring the objects and delivering it to the respective goal locations subject to task and motion constraints. For example, consider a robot in an office environment where it needs to deliver documents for evaluation to the respective project managers. At the task level, it is required that the robot first identifies the project in order to navigate to the respective sections, collect the documents and then deliver them to the project manager. A task planner computes a plan in terms of these symbolic actions, subject to minimizing a certain metric. This metric, for example, might correspond to different types of action costs or the number of actions. Since we are concerned with navigation, in this paper we associate the symbolic actions to their associated motion costs. Certain symbolic actions may not require robot motions. For example, for collecting a document, the robot may have to stay at a particular location for a given amount of time waiting for a human to place the document. Such actions are assigned a fixed cost}. Selecting the best set of discrete actions for a given objective requires computing the navigation costs \revtext{(and other fixed costs)} for each of these actions. Hence motion planning should be interleaved with task planning to compute the motion costs for each of the \revtext{respective} discrete actions. Though it can be argued that the motion costs can be approximated a priori and fed to the task planner, in large knowledge-intensive domains such an assumption can be harder to justify, especially in the presence of localization and map uncertainty. \revtext{Moreover}, real-world scenarios often induce uncertainties. Such uncertainties arise due to insufficient knowledge about the environment, inexact robot motion or imperfect sensing. In such scenarios, the robot poses or other variables of interest can only be dealt with, in terms of probabilities. Planning is therefore done in the \textit{belief} space, which corresponds to the probability distributions over possible robot states. Consequently, for efficient planning and decision making, it is required to reason about future belief distributions due to candidate actions and the corresponding expected observations. Such a problem falls under the category of Partially Observable Markov Decision Processes (POMDPs)~\cite{kaelbling1998AI}. Our motion planner is therefore equipped to perform planning in partially-observable state-spaces with motion and sensing uncertainty.
\begin{figure}[t!]
	\centering
		\includegraphics[scale=0.45]{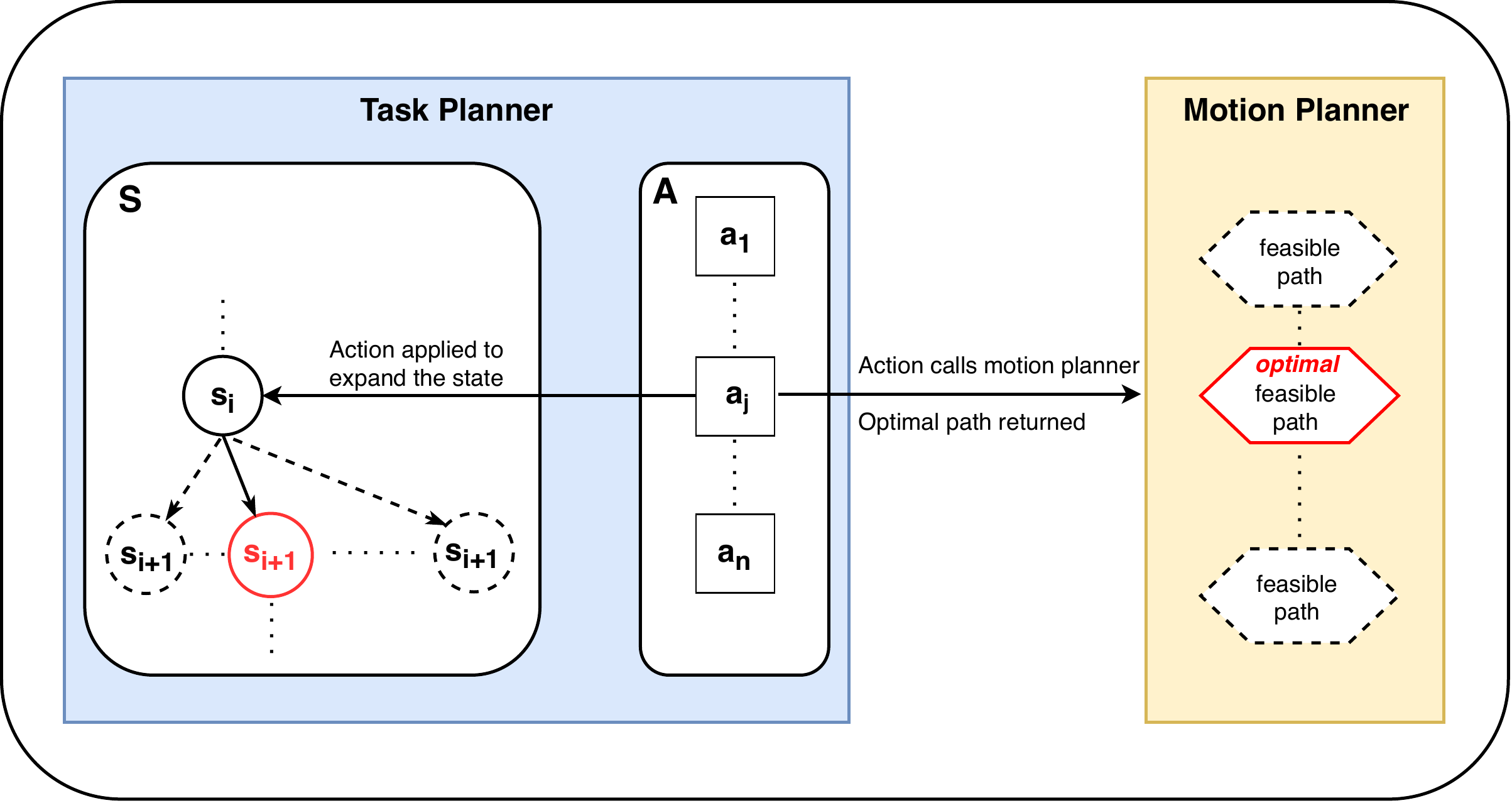}
		\caption{The discrete actions available to the planner are denoted by $A = \{a_1, a_2, a_3, \ldots, a_n\}$. Different motion plans are generated for the action that requires appropriate robot motion via an external module. This module is essentially a motion planner. The optimal path among the feasible motion plans is then selected, returning the optimal cost to the task planner. The corresponding action and the optimal path is the task-motion plan for changing the task state of the robot from $s_i$ to $s_{i+1}$.}
	\label{fig:sofar}
\end{figure}

This paper extends the work presented in~\cite{thomas2019ISRR} providing a comprehensive discussion and an extensive performance evaluation. Specifically, this paper contributes to the literature with a Motion-Planning-aware Task Planning (MPTP) approach providing an interface between task and motion planning for navigating in large knowledge-intensive domains. Such domains require a robot to reason about different objects and locations to navigate to, subject to minimizing (or maximizing) the navigation cost (objective function). Our task-motion interface layer facilitates this reasoning by communicating the motion feasibility and the corresponding planned motion costs to the task planner, synthesizing an optimal plan. To this end, we develop a probabilistically complete Task-Motion Planning (TMP) framework for mobile robot navigation under partial-observability, embedding a motion planner within a task planner through an interface layer. We would like to stress the fact that our implementation is independent of any particular form of cost function. In this paper, we use a standard cost function (see Section~\ref{sec:approach}) as the MPTP cost and compare it with different cost functions in Section~\ref{sec:results}.

An overview of our MPTP approach is shown in Fig.~\ref{fig:sofar}. We define $A = \{a_1,...,a_n\}$ as the finite set of symbolic/discrete actions available to the task planner. \revtext{For example, let us again consider an office setting where a robot is tasked with collecting and delivering documents. In such a setting, some of the actions include, \texttt{collect\_document}-- which might correspond to a human placing the document on the robot and therefore the robot waiting at a specific location for a certain duration, \texttt{deliver\_document}-- similar to \texttt{collect\_document} action but a human picks up the document, \texttt{goto\_region}-- corresponds to navigating through the environment}. Once an action that require appropriate robot motions to be generated is expanded by the task planner, a call to an external library is triggered. The symbolic parameters are then converted to their corresponding geometric instantiations. For example, for an action that takes the robot to a particular \revtext{cubicle/region}, the instantiations would be the different sampled poses in that cubicle. Once the map of the environment is obtained, the geometric instantiations can be pre-sampled. The instantiations give rise to different motion plans and the best among them is chosen according to a certain metric. The cost of the selected motion plan cost is then returned to the task planner as the cost of the corresponding action. The task-motion plan for changing the task state of the robot from the state $s_i$ to $s_{i+1}$ is the ordered tuple of the action $a_i$ and the corresponding optimal path. \revtext{For instance, in the office setting where a robot navigates from one cubicle ($s_i$) to another ($s_{i+1}$), the tuple is $\{\texttt{goto\_region}, \tau_i\}$. Here, \texttt{goto\_region} is the task-level action $a_i$ and $\tau_i$ is the planned trajectory for achieving this high-level action}. This tuple is appended for all the task-level actions to generate the complete task-motion plan. While our approach is applicable to any domain that require task-motion interaction, we establish the key ideas in Section~\ref{sec:approach} through two different navigation domains and further validate our approach in Section~\ref{sec:results} using the same.



%% file: related_work.tex

TMP has emerged as an active research area in the recent past, with particular focus on robot-based manipulation. Manipulation tasks are often rendered infeasible due to the  end-effector's reachability workspace. This calls for an integrated TMP approach to ensure geometric feasibility of high-level tasks.   

The genesis of TMP can be credited to Fikes and Nilsson for their work on STRIPS~\cite{fikes1971strips} which further led to the Shakey project~\cite{nilsson1984shakey}. Initial works on TMP performed task planning first, synthesizing a sequence of actions to be executed later by a motion planner. Shakey's planner performed a logical search first, assuming that the resulting robot motion plans can be formulated. This assumption limits the capability of the robot as the high-level actions may turn out to be non executable due to geometric limitations of the environment or the robot or both.~\cite{dornhege2009SSRR} interleaves task and motion planning by checking individual high-level action feasibility using \textit{semantic attachments}.~\cite{cambon2009IJRR} perform a combined search in the logical and geometric spaces using a state composed of both the symbolic and geometric paths. The aSyMov planner described in~\cite{cambon2009IJRR} adopts a combination of Metric-FF~\cite{hoffmann2003JAIR} and a sampling-based motion planner. In contrast, we use a temporal task planner, POPF-TIF~\cite{piacentini2015AI} with roadmap-based sampling, incorporating robot state uncertainty. Srivastava \textit{et al.}~\cite{srivastava2014ICRA} implicitly incorporate geometric variables, performing symbolic-geometric mapping using a planner-independent interface layer. \revtext{Erdem \textit{et al.}\cite{erdem2011ICRA} leverage a boolean
satisfiability (SAT) solver, computing a task-level plan and then refining it until a feasible motion plan is found.}

Kaelbling and Lozano-P\'{e}res~\cite{kaelbling2012aTR} propose a hierarchical approach that tightly integrates logical and geometric planning. The complexities arising out of long-horizon\footnote{Large environments require a robot to perform many actions to reach the goal, resulting in a long planning horizon\cite{kurniawati2011IJRR}.} planning are tackled to the extent that planning is done at different levels of abstraction, thereby reducing the long-horizons to a number of feasible sub-plans of shorter horizon. This regression\footnote{Goal regression is the process of planning backwards from the goal~\cite{ghallab2016book}.}-based planner assumes that the actions are reversible while backtracking. This work is extended in~\cite{kaelbling2013IJRR} to consider the current state uncertainty, modeling the planning problem in the belief space. \revtext{The hierarchical approach is also employed in~\cite{pandey2012BIOROB,de2013RSSws} to compute discrete actions with unbounded continuous variables. A geometric backtrack search is used to instantiate the symbolic actions in~\cite{lagriffoul2014IJRR}. They also prune certain geometric instantiations, reducing the complexity}. FFRob~\cite{garrett2018IJRR} performs task planning by performing search over a sampled finite set of poses, grasps and configurations. The authors of~\cite{garrett2018IJRR} extend the FF heuristics, incorporating geometric and kinematic planning constraints that provide a tight estimate of the distance to the goal. \revtext{Our approach is similar to FFRob in the sense that we also pre-sample robot configurations and then plans with them, incorporating motion constraints.}



Toussaint~\cite{toussaint2015IJCAI} performs optimization over an objective function based on the final geometric configuration (and the cost thereby), finding approximately locally optimal solutions by minimizing the objective function. The planning problem is modeled as a constraint satisfaction problem with symbolic states used to define the constraints in the optimization. \revtext{This \textit{logic-geometric programming} is applied to a four manipulator setting in~\cite{toussaint2017ICRA}}. Lozano-P\'{e}res and Kaelbling~\cite{lozano2014IROS} model the motion planning as a constraint satisfaction problem over a subset of the configuration space. Iteratively Deepened Task and Motion Planning (IDTMP) is a constraint-based task planning approach that incorporates geometric information to account for the motion feasibility at the task planning level~\cite{dantam2018IJRR}. In our architecture, the motion costs are returned to the task planner, similar to the motion planner information that guides the IDTMP task planner. IDTMP performs task-motion interaction using abstraction and refinement functions whereas we use \textit{semantic attachments}~\cite{dornhege2009ICAPS}.

Though the approaches discussed above fall under the category of TMP for manipulation, the scope of TMP is not limited to manipulation problems alone. TMP for navigation is pervasive in most real world scenarios. \revtext{For example, a mobile office robot may be tasked with collecting documents and delivering them across multiple floors.} Yet, TMP for robot navigation has received less attention in the past. Real-world planning problems in large scale environments often require solving several sub-problems. For example, while navigating to a goal, the robot might have to visit other places of interests. Visiting these places of interest are high-level tasks that can be addressed using traditional task planners. Yet, these symbolic planners cannot compute the exact motion costs for these tasks, let alone perform navigation and path planning. This calls for task plans that are motion planning aware, in terms of motion costs and its feasibility. 

\revtext{Task planning for robot Navigation Among Movable Obstacles (NAMO) is introduced in~\cite{stilman2008IJRR}, where each object is displace at most once throughout the plan. Van Den Berg \textit{et al.}~\cite{van2009WAFR} provide a probabilistically complete algorithm for the NAMO class of problems. However, the robot state is assumed to be known perfectly. In contrast, we plan in the belief space, computing an estimate of the robot state at each instant. Hauser and Latombe~\cite{hauser2009ICAPSws,hauser2010IJRR} consider \textit{multi-model motion planning} for manipulation and legged locomotion, wherein the space of feasible configurations consists of intersecting spaces of different dimensions. In~\cite{khandelwal2017IJRR} a TMP approach is presented in the context of Human-Robot Interaction (HRI). They integrate probabilistic reasoning with symbolic reasoning by implementing a spoken dialog system, enabling the robots to ask intelligent queries. Their task planner is based on Answer Set Programming (ASP)~\cite{lifschitz2002AI}. Jiang \textit{et al.}~\cite{jiang2019FITEE} focus exclusively on task planning in robotics, assuming that a feasible motion plan exists for the synthesized task plan. They provide a comparison between ASP-based and PDDL-based task planners using different benchmark domains and conclude that PDDL-based planners perform better on tasks with
long solutions, and ASP-based planners tend to perform better on shorter tasks. In this paper, we employ a PDDL-based task planner}. UP2TA~\cite{munoz2016RAS} develops a unified path planning and task planning framework for mobile robot navigation. In this approach, the robot is required to perform a series of tasks at different locations before returning back to the initial location. An interesting feature of UP2TA is its task planner heuristic, which is a combination of the FF heuristic~\cite{hoffmann2003JAIR} and the Euclidean distance between the waypoints associated with locations. The path planning layer computes the optimal path between each waypoint with the help of a Digital Terrain Model (DTM). Wong \textit{et al.}~\cite{wong2018optimal} develop a task planning approach that takes into account the optimal traversal costs\footnote{The costs are defined in terms of mechanical work and the objective is to find the path with optimal mechanical work. For more details, refer \revtext{to}~\cite{wong2018optimal}.} to synthesize a plan. Similar to UP2TA, they define tasks that are to be performed at different waypoints. However, the path planner pre-computes an optimal path for all pairs of waypoints, which are then passed to the task planner to find the optimal sequence of tasks. In contrast, we consider a general approach where the robot has to reason at a high-level about different objects or locations or regions to navigate to. The objects/locations/regions are instantiated to their geometric counterpart, by considering a set of sampled poses. For example, if a robot has to reach a location close to a chair, the geometric instantiations of this symbolic goal would correspond to a set of poses around the chair. 

\revtext{Jiang \textit{et al.}~\cite{jiang2019IROS} introduced a framework that integrates TMP with reinforcement learning that is robust to changes in the environment. The inner loop of their dual layer architecture is a TMP planner that generates task-motion plans to be sent to the outer loop. The outer loop executes the generated plans to learn from rewards. In contrast MPTP is a purely planning approach. Lo \textit{et al.}~\cite{lo2018AAMAS} introduced PETLON, a purely planning approach for navigation that is task-level optimal and is the work closest to our approach. The inner loop in~\cite{jiang2019IROS} uses a TMP planner that is similar to PETLON}. However, \revtext{in PETLON,} the action costs returned by the motion planner is the trajectory length and complete observability is assumed. In contrast, our framework is more general, since we additionally consider the cost due to \revtext{motion and sensing} uncertainty and the distance to the goal. It is to be noted that our approach is not limited to any particular cost function and can be easily adapted to support any general \revtext{cost} formulation. In Section~\ref{sec:results}, we benchmark the scalability of our approach \revtext{and provide a comparison with PETLON by considering a motion planner that evaluates the geometric-level cost of navigation. In this way we compare MPTP to PETLON by adapting our cost function to incorporate only the geometric-level cost of traversing from one location to another}. \revtext{Further, PETLON first compute a task plan using an admissible heuristic which is then sent to the motion planner for cost evaluation. This updates the heuristic and a refinement process repeats until the optimal plan is found. In contrast, MPTP evaluates the motion cost as each action is expanded by the task planner and hence the plan returned is optimal and needs no refinement}.

%% file: preliminaries.tex
We begin by formally defining the notions of task and motion planning. Then, we state the TMP problem that we discuss in this paper. The notations and formalism correspond to that of a state-transition system~\cite{ghallab2016book}.

\subsection{Task Planning}
Task planning or classical planning can be defined as the process of finding a discrete sequence of actions from the current state to a desired goal state~\cite{ghallab2016book}.

\begin{defn}A \textit{task} domain $\Omega$ can be represented as a state transition system and is a tuple $\Omega = (S, A, \gamma, s_0, S_g)$ where:
\label{def:one}
\end{defn}
\begin{itemize}
\item $S$ is a finite set of states, each state is a conjunction of propositions\footnote{A proposition is represented by a tuple of elements, which may be constants or variables, and can be negated~\cite{bylander1994AI}.};
\item $A$ is a finite set of actions;
\item $\gamma : S \times A \rightarrow S$ is the state transition function such that $s' = \gamma(s, a)$;
\item $s_0 \in S$ is the start state;
\item $S_g \subseteq S$ is the set of goal states.
\end{itemize}

\begin{defn} The task \textit{plan} for a task domain $\Omega$ is the sequence of actions $a_0,...,a_n$ such that $s_{i+1} = \gamma(s_i, a_i)$, for $i = 0,...,n$ and $s_{n+1}$ \textit{satisfies} $S_g$.
\end{defn}
 
The Planning Domain Definition Language (PDDL)~\cite{mcdermott1998AIPS} being the de facto standard syntax for task planning, we resort to the same for modeling our task domain. PDDL is an action-centred language, where each action $a_i$ is described as a tuple $a_i = (pre_{a_i},eff_{a_i})$, where $pre_{a_i}$ (a set of preconditions for $a_i$ ) is a conjunction of propositions with either positive or negative terms that must hold for action execution and $eff_{a_i}$ (the set of effects of $a_i$) is a conjunction of positive ($eff^+_{a_i}$) and negative ($eff^-_{a_i}$) propositions that are added or deleted upon action application, thereby changing the system state. The set of positive effects $eff^+_{a_i}$ contains propositions that become true upon the execution of action $a_i$ and the set of negative effects $eff^-_{a_i}$ contains propositions that evaluates to false upon action execution. An action $a_i$ is said to be applicable to a state $s_i$ if each proposition of the preconditions \revtext{holds} in $s_i$, that is, $pre_a \subseteq s_i$.  If an action $a_i$ is applicable in state $s_i$, the corresponding successor state $s_{i+1}$ is obtained as, $s_{i+1} = \gamma(s_i, a_i)$, where $s_{i+1} = (s_i \setminus eff^-_{a_i}) \cup eff^+_{a_i}$. A valid plan is a sequence of actions that when executed from $s_0$ results in $S_g$. 

A planning problem with PDDL is created by providing a domain description that describes the predicates and action schemas with free variables, and a problem description that specifies the objects, initial state and the goal condition. The objects are used to instantiate the predicates and action schemas, through a process called grounding. Grounding is the process by which every combination of objects is used to replace the free variables in predicates and action schemas to obtain propositions and ground actions respectively. In this paper, we use an extension of PDDL~\cite{fox2003JAIR} that supports durative actions and numeric-valued fluents. Temporal planning introduces the possibility of computing concurrent plans. A temporal task domain can be defined by extending the task domain in Definition~\ref{def:one} as follows 

\begin{defn}A \textit{temporal} task domain $\Omega$ can be represented as state transition system and is a tuple $\Omega = (S, A, \gamma, s_0, S_g)$ where:
\end{defn}
\begin{itemize}
\item $S$ is a finite set of states;
\item $V$ is a set of real valued variables;
\item $A$ is a finite set of actions;
\item $\gamma : S \times A \rightarrow S$ is the state transition function such that $s' = \gamma(s, a)$;
\item $s_0 \in S \cup V$ is the start state;
\item $S_g \subseteq S \cup V$ is the set of goal states.
\end{itemize}

A durative action is a tuple  $a_i = (pre_{a_i},eff_{a_i},dur_{a_i})$, where $pre_{a_i}$ and $eff_{a_i}$ are temporally annotated by specifying conditions/effects that holds at the \textit{start}, \textit{end} or during the \textit{entire} action interval and are expressed using the constructs \textit{at start}, \textit{at end} and \textit{over all} respectively. Note that these constructs are specific to PDDL formalism. $dur_{a_i}$ corresponds to the duration of action $a_i$.

\subsection{Motion Planning}
Motion planning finds a sequence of collision free poses from a given initial/start pose (position and orientation) to a desired goal pose~\cite{latombe1991robot}.

\begin{defn}A \textit{motion planning problem} is a tuple $M = (C, f, q_0, G)$ where:
\end{defn}
\begin{itemize}
\item $C$ is the configuration space or the space of possible robot poses;
\item $f =\{0,1\}$ determines if a configuration/pose is in collision ($f=0$) or not ($C_{free}$ with $f =1$). $C_{free}$ denotes the set of all poses that are not in collision;  
\item $q_0$ is the initial configuration;
\item $G$ is the set of goal configurations.
\end{itemize}

\begin{defn} A motion \textit{plan} for $M$ finds a valid trajectory in $C$ from $q_0$ to $q_n \in G$ such that $f$ evaluates to true for $q_0,...,q_n$.
\end{defn}

In addition to the sequential form of the definition above, a motion plan can also be defined by a continuous trajectory

\begin{defn} A motion \textit{plan} for $M$ is a function of the form $\tau : [0, 1] \rightarrow C_{free}$ such that $\tau(0) = q_0$ and $\tau(1) \in G$. 
\end{defn}
We will use a combination of the two to define the TMP problem and use roadmap based motion planner to generate collision free configurations.  

\subsection{Task-Motion Planning}
TMP essentially involves combining discrete and continuous decision-making to facilitate efficient interaction between the two domains. Starting from an initial state, TMP synthesizes a plan to a goal state by a concurrent or interleaved set of discrete actions and continuous collision-free motions. Below we define the TMP problem formally.

\begin{defn}A \textit{task-motion planning} is a tuple $\Psi =(C, \Omega, \phi, \xi, q_0)$ where:
\end{defn}
\begin{itemize}
\item $\phi : S  \rightarrow 2^ C$, is a function mapping states to the configuration space. \revtext{For example, if $s$ represents the task state--- the robot is in a corridor, then $\phi(s)$ corresponds to all configurations such that the robot is in the corridor}; 
\item $\xi : A  \rightarrow 2^ C$, is a function mapping actions to motion plans. \revtext{We recall here that motion planning is essentially computing collision free poses in $C$.}
\end{itemize}

\begin{defn}The \textit{TMP problem} for the TMP domain $\Psi$ is to find a sequence of actions $a_0,...,a_n$ such that $s_{i+1} = \gamma(s_i, a_i)$, $s_{n+1} \in S_g$ and to find a sequence of motion plans $\tau_0,...,\tau_n$ such that for $i = 0,...,n$, it holds that
\end{defn}

\vspace{-0.8cm}

\begin{align}
& \tau_i(0) \in \phi(s_i) \ \textrm{and} \ \tau_i(1)  \in \phi(s_{i+1})  \\
&\tau_{i+1}(0) = \tau_i(1)   \\
&\tau_i \in \xi(a_i)
\end{align}

\subsection{Problem Definition}
In this paper, we consider the TMP problem for a mobile robot operating in a partially-observable environment. The map of the environment is either known a priori or is built using a standard Simultaneous Localization and Mapping (SLAM) algorithm\footnote{http://wiki.ros.org/slam\_gmapping/}. At any time $k$, we denote the robot pose (or configuration $q_k$) by $x_k\doteq(x, y, \theta)$, the acquired measurement is denoted by $z_k$ and the applied control action is denoted as $u_k$. We consider a standard motion model with Gaussian noise



\vspace{-0.40cm}
\begin{equation}
x_{k+1} = f(x_k,u_k,w_k) \  ,  \ w_k \sim \mathcal{N}(0,W_k)
\label{eq:odometry_model}
\end{equation}

\noindent
where $w_k$ is the random unobservable noise, modeled as a zero mean Gaussian. To process the landmarks in the environment we measure the range and the bearing of each landmark relative to the robot's local coordinate frame. In general, we consider the observation model with Gaussian noise 

\vspace{-0.35cm}
\begin{equation}
z_k = h(x_k) + v_k \  ,  \ v_k \sim \mathcal{N}(0,Q_k)
\label{eq:measurement_model}
\end{equation}
 

It is to be noted that we assume data association as solved and hence given a measurement we know the corresponding landmark that generated it. This is not a limitation and our approach can be extended to incorporate reasoning regarding data association, as shown recently in~\cite{pathak2018IJRR}. The motion (\ref{eq:odometry_model}) and observation (\ref{eq:measurement_model}) models can be written probabilistically as
$p(x_{k+1}|x_k, u_k)$ and $p(z_k|x_k)$ respectively. Given an initial distribution $p(x_0)$, and the motion and observation models, the posterior probability distribution at time $k$ can be written as

\vspace{-0.2cm}
\begin{equation}
p(X_{0:k}|Z_{0:k},U_{0:k-1}) = p(x_0)\prod_{i=1}^k p(x_{k}|x_{k-1}, u_{k-1}) p(z_k|x_k)
\end{equation}
where $X_{0:k} \doteq \{x_0,...,x_k\}$, $Z_{0:k}  \doteq\{z_0,...,z_k\}$ and $U_{0:k-1} \doteq \{u_0,...,u_{k-1}\}$. This posterior probability distribution is the \textit{belief} at time $k$, denoted by $b[X_k] \sim \mathcal{N} (\mu_k, \Sigma_k)$. Similarly, given an action $u_k$, the propagated belief can be written as

\vspace{-0.3cm}
\begin{equation}
b[\bar{X_{k+1}}] = p(X_{0:k}|Z_{0:k}, U_{0:k-1})p(x_{k+1}|x_k, u_k)
\end{equation}

Given the current belief $b[X_k]$ and the control $u_k$, the propagated belief parameters can be computed using the standard Extended Kalman Filter (EKF)\revtext{~\cite{kalman1960ASME}} prediction as 

\vspace{-0.5cm}
\begin{equation}
\begin{split}
\bar{\mu}_{k+1} & = f(\mu_k, u_k)\\
\bar{\Sigma}_{k+1}   & = F_{k} \Sigma_k F_{k}^T + V_kW_kV_k^T
\end{split}
\label{eq:predict}
\end{equation}
where $F_k$ is the Jacobian of $f(\cdot)$ with respect to $x_k$ and $V_k$ is the Jacobian of $f(\cdot)$ with respect to $u_k$. For brevity, the linearized process noise will be denoted as $R_k = V_kW_kV_k^T$. Upon receiving a measurement $z_k$, the posterior belief $b[X_{k+1}]$ is computed using the EKF update equations

\vspace{-0.4cm}
\begin{equation}
\begin{split}
K_k     & = \bar{\Sigma}_{k+1} H_k^T(H_k \bar{\Sigma}_{k+1}  H_k^T + Q_k)^{-1}\\
\mu_{k+1} & = \bar{\mu}_{k+1} + K_k(z_{k+1}-h(\bar{\mu}_{k+1}))\\
\Sigma_{k+1} & = (I -K_k H_k)\bar{\Sigma}_{k+1} 
\end{split}
\label{eq:update}
\end{equation}

\noindent
where $H_k$ is the Jacobian of $h(\cdot)$ with respect to $x_k$, $K_k$ is the Kalman gain and $I \in \mathbb{R}^{3 \times 3}$ is the identity matrix.

%% file: methodology.tex
PDDL-based planning frameworks are limited, as they are incapable of handling rigorous numerical calculations\footnote{PDDL+~\cite{fox2006JAIR}, an extension of PDDL supports mixed discrete and continuous non-linear changes.}. Most approaches perform such calculations via external modules or \textit{semantic attachments}, e.g.~\cite{dornhege2009ICAPS}. The term semantic attachment was coined by Weyhrauch~\cite{weyhrauch1980AI} to describe the association of algorithms to function and predicate symbols via external procedures. However, the effects returned by these semantic attachments are not exploited in identifying \textit{helpful actions} during search and hence do not provide any heuristic guidance, deeming the task unsolvable most often~\cite{bernardini2017ICAPS}. An action is considered \textit{helpful} if it achieves at least one of the lowest level goals in the relaxed plan to the state at hand~\cite{hoffmann2003JAIR}. Recently, Bernardini \textit{et al.}~\cite{bernardini2017ICAPS} developed a PDDL-based temporal planner to implicitly trigger such external calls via a specialized semantic attachments called \textit{external advisors}. They classify variables into direct ($V^{dir}$), indirect ($V^{ind}$) and free ($V^{free}$). $V^{dir}$ and $V^{free}$ variables are the normal PDDL function variables whose values are changed in the action effects, in accordance with PDDL semantics. $V^{ind}$ variables are affected by the changes in the $V^{dir}$ variables. A change in a $V^{dir}$ variable invokes the external advisor which in turn computes the $V^{ind}$ variables. The Temporal Relaxed Plan Graph (TRPG)~\cite{coles2010ICAPS} construction stage of the planner incorporates the indirect variable values for heuristic calculation, thereby synthesizing an efficient goal-directed search. We employ this semantic attachment based approach for the task-motion interface. The overall procedure and the interface layer are discussed in detail in the remainder of this Section. 


\begin{figure}[]
	\centering
		\includegraphics[scale=0.25]{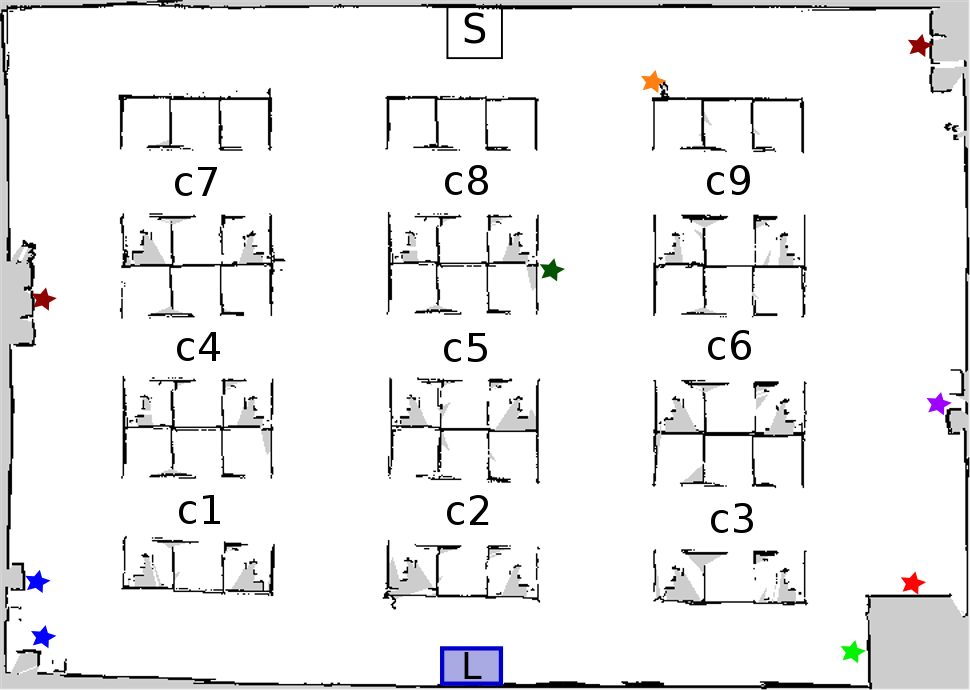}
		\caption{\revtext{Map of the office environment obtained after a SLAM session.}}
	\label{fig:office}
\end{figure}

\subsection{Task Planning} 

TMP for navigation requires that the task planner takes into account the motion feasibility and the corresponding motion costs while synthesizing a plan. As opposed to the manipulation domain, where the motion feasibility is corroborated with the end-effector's reachability workspace, in navigation domains this is often validated against the cost constraints, for example, a robot navigating in a corridor with a bound on the pose covariance to avoid collisions. As such, any task planner customized to enable the task-motion interface can be employed for our approach. In our tests, PDDL is used to define the task domain. 

Below, we elucidate the PDDL formalism for two different navigation domains that we have considered. It is to be noted that the semantic attachment procedure is domain independent and remains the same in both the domains. But the PDDL domain and problem description differ, as the two domains are different in nature. In the first domain, the underlying roadmap for motion planning does not change during plan computation. However, in the second domain, the roadmap is updated during plan computation. Description of the two domains are detailed below. 

\subsubsection{Office Domain}
\label{sec:nav_domain1}
\newsavebox{\newlisting}
 \lstset{basicstyle=\small}
 \lstset{escapeinside={<@}{@>}} 
\begin{lrbox}{\newlisting}
\begin{lstlisting}

(<@\textcolor{cyan}{:durative-action}@> goto_region
 <@\textcolor{olive}{:parameters}@> (?v - robot ?from ?to - region)
 <@\textcolor{olive}{:duration}@> (= ?duration 100)
 <@\textcolor{olive}{:condition}@> (at start (robot_in ?v ?from))
 <@\textcolor{olive}{:effect}@> (and (at start (not (robot_in ?v ?from))) 
 (at start (increase (triggered ?from ?to) 1))
 (at end (robot_in ?v ?to)) (at end (assign (triggered ?from ?to) 0))	
 (at end (increase (act-cost) (external)))
 (at end (increase (goal-trace) (bound))))

(<@\textcolor{cyan}{:durative-action}@> collect_document
 <@\textcolor{olive}{:parameters}@> (?v - robot ?r - region)
 <@\textcolor{olive}{:duration}@> (= ?duration 20)
 <@\textcolor{olive}{:condition}@> (and (at start (robot_in ?v ?r)) (at start (> (get ?r) 0))
 (over all (robot_in ?v ?r)))
 <@\textcolor{olive}{:effect}@> (and (at end (collected ?r))(at end (increase (act-cost) 4))))

(<@\textcolor{cyan}{:durative-action}@> goto_lift
 <@\textcolor{olive}{:parameters}@> (?v - robot ?from ?to - region)
 <@\textcolor{olive}{:duration}@> (= ?duration 100)
 <@\textcolor{olive}{:condition}@>(at start (robot_in ?v ?from))
 <@\textcolor{olive}{:effect}@> (and (at start (not (robot_in ?v ?from))) 
 (at start (increase (triggered ?from ?to) 1))
 (at end (reached ?to)) (at end (assign (triggered ?from ?to) 0))	
 (at end (increase (act-cost) (external))))

\end{lstlisting}
\end{lrbox}

\begin{figure}[]

 \scalebox{0.93}{\usebox{\newlisting}}\hfill%
\caption{A fragment of the PDDL office domain.}
 \label{fig:domain}
\end{figure}

We consider a robot navigating in an office environment \revtext{to collect and deliver documents. The map of the environment following a SLAM session is shown in Fig~\ref{fig:office} (snapshot of the environment can be seen in Fig.~\ref{fig:gazebo})}. The regions $c_1,\ldots,c_9$ are cubicles and $L$ denotes a lift. The robot, starting from region $S$ has to visit certain cubicles to receive documents. Navigating to cubicles/regions is encoded using a single high-level action \texttt{goto\_region}. Once a robot reaches a cubicle from which a document is to be collected, we assume that a human places the requisite document. \revtext{Thus, the robot needs to wait at the specific location for a fixed duration of time in which the human places the required document on the robot}. This is encoded using a high-level action \texttt{collect\_document}. These documents then have to be delivered to another floor, which implies using the lift $L$. Navigating to the lift is modeled using a different high-level action \texttt{goto\_lift}. This is because, unlike the action \texttt{goto\_region}, \texttt{goto\_lift} is to be performed only if the robot has collected all the necessary documents to be delivered. The stars with different colors represent certain unique features assumed to be known and modeled like, printer, trash can, lounge, that aids the robot in better localization. Hence, once the robot knows the regions to visit, then it suffices to perform \texttt{goto\_region} actions and collect the documents from these regions. However, to synthesize an optimal plan it is necessary to sequence these actions in an order that minimizes the cost function. It is therefore inevitable to obtain the motion costs of these \texttt{goto\_region} actions, so as to accurately synthesize the optimal plan.

A fragment of the PDDL domain is shown in Fig.~\ref{fig:domain}. \revtext{The PDDL domain dynamics is specified through a set of durative actions (\texttt{:durative-action}). We use the following parameters to model these actions: \texttt{?v} is the name of the robot, \texttt{?from} is the cubicle the robot is currently at and \texttt{?to} is the cubicle to which the robot needs to move, \texttt{?r} corresponds to the different regions or cubicles in the environment. Each action is described using \texttt{:condition} and \texttt{:effect}, as defined in Section~\ref{sec:preliminaries}, and defines the conditions and effects that holds at the start (\texttt{at start}), end (\texttt{at end}) or during the entire action interval (\texttt{overall}), respectively. The predicate \texttt{robot\_in} checks if the robot is in a particular region. The function \texttt{triggered} encodes the fact that the robot is moving from one cubicle (\texttt{from}) to another (\texttt{to}). The functions \texttt{get} and \texttt{collected} model the cubicles from which the document is collected and whether it has been collected. Finally, \texttt{act-cost} stores the cost associated with the actions and \texttt{goal-trace} keeps the robot state uncertainty bounded}. The actions \texttt{goto\_region} and \texttt{goto\_lift} invoke the external module call once the facts \texttt{(increase (act-cost) (external))} and \texttt{(increase (goal-trace) (bound))} are encountered. Here, \texttt{act-cost}, \texttt{goal-trace} are the direct variables in $V^{dir}$ and \texttt{external}, \texttt{bound} are the indirect variables $V^{ind}$. The function \texttt{(triggered ?from ?to)} is assigned the numerical value $1$ each time the actions are expanded and re-initialized to $0$ once the action duration is completed. In this way, the grounded variables \texttt{from} (start) and \texttt{to} (goal) are communicated to the motion planner. The variables \texttt{external} and \texttt{bound} returns the motion cost and the goal covariance trace respectively, which are computed by the external module. The action \texttt{collect\_document} does not invoke the motion planner. In the problem description, the function \texttt{(get ?r)}, where \texttt{r} is a free variable denoting cubicles, is initialized to $1$ for the cubicles from which the documents are to be collected and to $0$ for the remaining.

\subsubsection{Corridor Domain}
\label{sec:nav_domain2}

\begin{figure}[t!]

  \subfloat[\revtext{\texttt{goto\_door}}]{\includegraphics[scale=0.28]{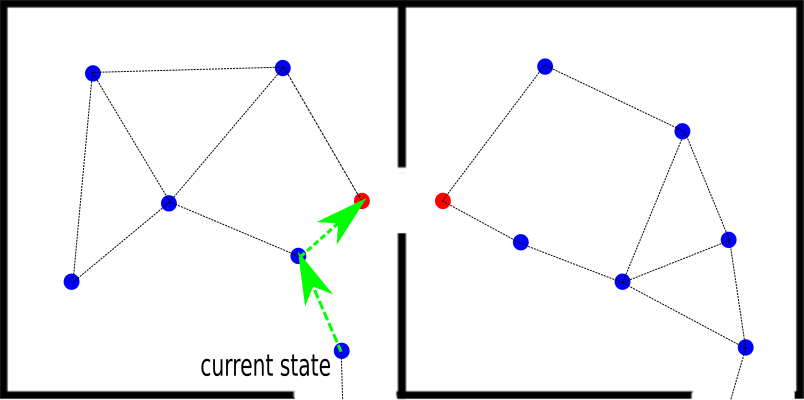}} \hspace{0.1cm}%
  \subfloat[\revtext{\texttt{goto\_door}}]{\includegraphics[scale=0.28]{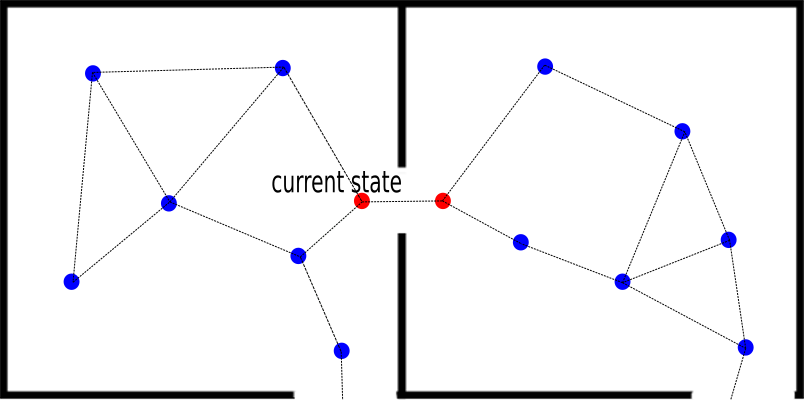}} \\
  \subfloat[\revtext{\texttt{goto\_room}}]{\includegraphics[scale=0.28]{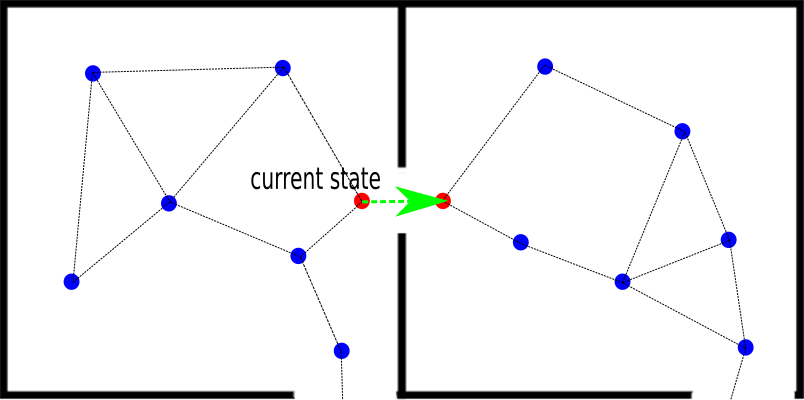}} \hspace{0.1cm}%
  \subfloat[\revtext{\texttt{goto\_room}}]{\includegraphics[scale=0.28]{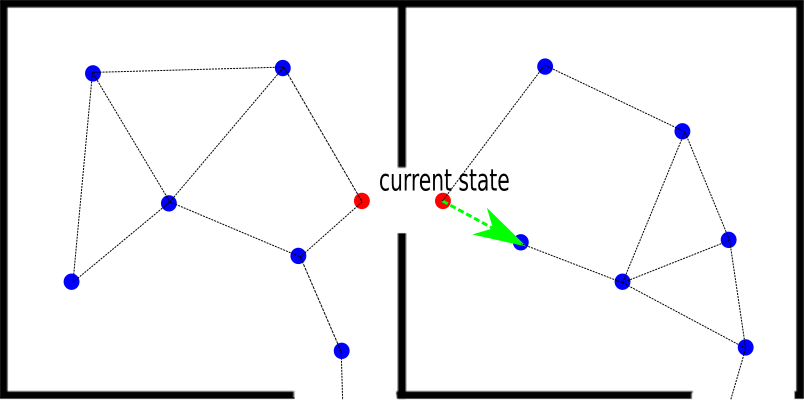}}
  \caption{The addition and deletion of an edge to the PRM graph. The red nodes are the ones that are close to the door. (a) Shows a possible path in green, when the \texttt{goto\_door} action is expanded. Note that there is no edge between the two red colored nodes. (b) Upon satisfying the trace constraint, an edge added between the two nodes close to the door. (c) The \texttt{goto\_room} action takes the robot to the next room. (d) As the robot \revtext{navigates towards the first node (red colored node) in the new room, the edge connecting it to the room from which the robot traversed is removed from the roadmap.}}
  \label{fig:edge}
\end{figure}

We consider a navigation domain, similar to the one in~\cite{jiang2019FITEE}, wherein a robot navigates through a building floor that consists of several rooms connected to one another through a corridor. These rooms have doors, which can either be closed or open, connecting them to the corridor. In addition, some of the rooms are also accessible from each other, through doors in between them. The robot can navigate through the entire building by opening these doors. We assume that once the robot is near to a closed door that directly connects a room to the corridor, a \textit{human opens} the door to allow the robot to pass through. Navigating to rooms can hence be encoded using a single high-level action \texttt{goto\_room}. However, the doors between \revtext{any two} rooms are automatic, that opens only when the robot is directly in front of the door. This requires the robot to navigate to the door and is encoded using the high-level action \texttt{goto\_door}. Upon reaching the goal, since the robot is uncertain about its pose, the robot can be anywhere within its current belief distribution. Taking this into account, on reaching the door it is open only if the trace of the pose covariance is within a certain bound $\eta$. If the trace is within the bound, an edge is added to the Probabilistic Roadmap (PRM)~\cite{kavraki1996IEEE} graph between the current node and the nearest node in the next room to which the robot can navigate via the door. Once the robot traverses the door to reach the next room, the newly added edge is removed from the roadmap. This process is illustrated in Fig.~\ref{fig:edge}. The addition and deletion of edges is performed by the external module.  

A fragment of the corridor PDDL domain is shown in Fig.~\ref{fig:domain2}. \revtext{Similar to the office domain, we use the following parameters: \texttt{?from} is the room the robot is currently at and \texttt{?to} is the room which the robot needs to visit, \texttt{?d} is any door. The predicate \texttt{visited\_in} checks if the robot has visited a room, \texttt{hasdoor} checks if the room has a door that opens to another room, and \texttt{expanded} model the change in the roadmap.} Similar to the previous domain, the actions \texttt{goto\_room} and \texttt{goto\_door} invoke the external module call once the fact \texttt{(increase (act-cost) (external))} is encountered. Here, \texttt{act-cost} is the direct variable in $V^{dir}$ variable and \texttt{external} is the indirect variable in $V^{ind}$. The function \texttt{(triggered ?from ?to)} and \texttt{(expanded ?r ?d)} are assigned the value of $1$ each time the actions are expanded and re-initialized to $0$ once the action duration is completed. This is performed so that the grounded variables \texttt{from} (start) and \texttt{to} (goal) as well as \texttt{r} (start) and \texttt{d} (goal) are communicated to the motion planner. The variables \texttt{from}, \texttt{to} and \texttt{r} are used to denote the rooms and the variable \texttt{d} represents the doors available. This can be seen in the \texttt{parameters} definition of the actions. The variable \texttt{external} returns the motion cost computed by the external module.   


\newsavebox{\newlist}
 \lstset{basicstyle=\small}
 \lstset{escapeinside={<@}{@>}} 
\begin{lrbox}{\newlisting}
\begin{lstlisting}

(<@\textcolor{cyan}{:durative-action}@> goto_room
 <@\textcolor{olive}{:parameters}@> (?from ?to - room)
 <@\textcolor{olive}{:duration}@> (= ?duration 100)
 <@\textcolor{olive}{:condition}@> (and (at start (robot_in ?from)) (at start                (connected ?from ?to)))
 <@\textcolor{olive}{:effect}@> (and (at start (not (robot_in ?from))) 
 (at start (increase (triggered ?from ?to) 1))
 (at end (robot_in ?to)) (at end (assign (triggered ?from ?to) 0))	
 (at end (increase (act-cost) (external))) (at end (visited ?to))))

(<@\textcolor{cyan}{:durative-action}@> goto_door
 <@\textcolor{olive}{:parameters}@> (?r - room ?d - door)
 <@\textcolor{olive}{:duration}@> (= ?duration 40)
 <@\textcolor{olive}{:condition}@>(and (at start (robot_in ?r)) (at start (hasdoor ?r ?d))
 (over all (robot_in ?r)))
 <@\textcolor{olive}{:effect}@> (and (at start (increase (expanded ?r ?d) 1))
 (at end (assign (expanded ?r ?d) 0))	
 (at end (increase (act-cost) (external)))))

\end{lstlisting}
\end{lrbox}

\begin{figure}[]

 \scalebox{0.93}{\usebox{\newlisting}}\hfill%
\caption{A fragment of the PDDL corridor domain.}
 \label{fig:domain2}
\end{figure}

\subsection{Motion Planning} 

Independently of the domain, we use a sampling based PRM to instantiate robot poses for the task actions. To begin with, the initial mean and covariance of the robot pose is assumed to be known. This means that the initial state $s_0$ corresponds to a single pose instantiation $q_0$. The regions to be navigated to are also instantiated into poses, by sampling from the pose space within each region. Once an action $a_i$ is expanded by the task planner, the corresponding start and goal states, that is $s_i$ and $s_{i+1}$ are communicated to the motion planner. This is facilitated by the functions \texttt{triggered} and \texttt{expanded}, as detailed in the previous \revtext{section}. For example, the task state $s_i$ might specify that the robot is in cubicle $c_2$ and the goal state $s_{i+1}$ can be for the robot to reach cubicle $c_4$. In this scenario $\phi(s_i)$ and $\phi(s_{i+1})$, that is, the mapping from states to configurations, correspond to all possible poses such that the robot is in cubicles $c_2$ and $c_4$ respectively. Since the set of possible poses is infinite, we randomly sample a set of poses corresponding to each task state $s_i$. It is to be noted that this sampling is an independent problem and this set is incorporated while building the entire roadmap. For each region $s_i$, the number of pose instantiations will be denoted by $s_i^n$ and a particular instantiation by $s_i^{n_k}$. With the pose instantiation of $s_i$ as the start node, for each pose instantiation of $s_{i+1}$, we simulate a sequence of controls along each edge starting from $s_i^{n_k}$ and ending in $s_{i+1}^{n_j}$, estimating the beliefs at the each of these nodes using~(\ref{eq:predict})-~(\ref{eq:update}). The $s_{i+1}^{n_j}$ that corresponds to the minimum cost is then selected as the goal pose to reach, for the state $s_{i+1}$. Thereafter, this instantiation becomes the start node when an expansion is attempted from state $s_{i+1}$. It is true that PRM is in the configuration space and not in the belief space, but the basic problem remains the same since we are essentially finding a sequence of actions that minimizes the objective function which is a function of the resulting beliefs. Our PRM approach is similar to \revtext{the Belief Roadmap (BRM)~\cite{prentice2009IJRR} approach} and differs in the way one-step belief updates are performed. Moreover, BRM assume maximum likelihood observations but we do not.

Since we plan in the belief space of the robot state, given the mean and covariance of the starting node we propagate the belief along the edges of the PRM as the roadmap is expanded during the search. Belief update is performed upon reaching a node if a landmark is successfully detected by the robot's perception system. Since we are in the planning phase and yet to obtain observations, we simulate future observations $z_{k+1}$ given the propagated belief $b[\bar{X_{k+1}}]$, the set of landmarks $L_{\mathbb{N}}={l_1,\ldots,l_n}$ and the measurement model (\ref{eq:measurement_model}). \revtext{In this work, we model landmarks using AprilTags~\cite{olson2011ICRA} which are placed on the objects of interest.} Given a pose $x \in b[\bar{X_{k+1}}]$, the nominal observation $\hat{z} = h(x, l_i)$ is corrupted with noise to obtain $z_{k+1}$, which is then used to compute the posterior belief. 
 
\subsection{Task-Motion Planning for Navigation}

\begin{algorithm}[]
\caption{TMP for Navigation in Belief Space}
\label{algo:TMP}
\begin{algorithmic}[1]
\Require{\revtext{$\Psi =(C, \Omega, \phi, \xi, q_0)$: Task-Motion domain}, \revtext{$\eta$: Uncertainty budget}}
\While{true}
\State{$a_i$ $\leftarrow$ task planning($\Omega$)}
\LineComment $a_i$ = an action selected to expand the next state
\State{$\pi^*$ $\leftarrow$ $\emptyset$ \qquad \textcolor{mygray}{// Task-Motion Plan} }
\If{$a_i$ $\in$ $A_s$}
\State{External module $\leftarrow$ $V^{dir}$}
\LineComment $V^{dir}$  $= \{act-cost, goal-trace\}$
\State{current task state $\leftarrow$ $s_i$, next task state $\leftarrow$ $s_{i+1}$}
\State {$c$ $\leftarrow$ $\emptyset$, T $\leftarrow$ $\emptyset$ }
\State{current task state $\leftarrow$ $\phi(s_i)$, next task state $\leftarrow$ $\phi(s_{i+1})$}
 \For{\textbf{each} $s_{i+1}^{n_j} \in \phi(s_{i+1})$}

\State{start node $\leftarrow$ $s_i^{n_k}$, goal node $ s_{i+1}^{n_j}$}
\State{Belief space search from start node to goal node.}

\State{$c$ $\leftarrow$ $c^j$, $T$ $\leftarrow$ $\tau_i^j$ }
   \EndFor

\State { $j^*$ = \textbf{arg\,min} $c$ }
\State {$\tau_i \leftarrow \tau_i^{j^*}$}
\LineComment $\tau_i$ is the selected motion plan to arrive at the task state $s_{i+1}$.

\State{$V^{ind}$ $\leftarrow$ External module}
\State{$\pi^*$ $\leftarrow$ append($\pi^*$, $(a_i,\tau_i)$)}

\EndIf

\EndWhile
\State \Return{$\pi^*$}
\end{algorithmic}
\end{algorithm}

\begin{algorithm}[h]
\caption{\revtext{Belief space search}}
\label{algo:bsp}
\revtext{
\begin{algorithmic}[1]

\Require{Roadmap (sampled poses and edges), start node $n$ with belief ($\mu_n,\Sigma_n$) corresponding to start state $s_i$, goal node ($\phi(s_{i+1})$)}
\State{$\tau_i$ $\leftarrow$ $n$ }
\While{$\phi(s_{i+1})$ not reached}
\For{\textbf{each} edge from $n$ to $n'$}

\State{Propagate the belief (\ref{eq:predict})}
\If{Landmark within sensing range}
\State{Compute posterior belief (\ref{eq:update}).}
\EndIf
\State{Select $n'$ with minimum cost.}
\State{$c$ $\leftarrow$ minimum cost, $\tau_i$ $\leftarrow$ append($\tau_i$,$n'$)}
\State{$n' = n$}
\EndFor
\EndWhile

\State \Return{$c, \tau_i$}
\end{algorithmic}}
\end{algorithm}
 In our approach, the interface between task and motion planning occurs through semantic attachments. Formally, semantic attachment can be defined as
\begin{defn} Semantic attachments is a functional mapping from the set of direct variables to the set of indirect variables, that is, $\chi: V^{dir} \rightarrow V^{ind}$.
\end{defn}

\revtext{We recall here that for the \textit{office domain} $V^{dir} = \{\texttt{act-cost}, \texttt{goal-trace}\}$ and $V^{ind} = \{\texttt{external}, \texttt{bound} \}$. For the \textit{corridor domain}, we have $V^{dir} = \{\texttt{act-cost}\}$ and $V^{ind} = \{\texttt{external} \}$}. The planner receives as input- the PDDL domain, problem description, the shared library and other input parameters. The input parameter specifies the regions/rooms and the corresponding pose instantiations. For the office domain, these pose instantiations are the poses that lie inside the cubicles  and for the corridor domain they are the poses that lie inside the rooms. These poses are sampled once the map of the environment is available as described in the previous \revtext{section}.

An overview of our TMP approach is presented in Algorithm~\ref{algo:TMP}. The external module computes the $V^{ind}$ values and is invoked only when a change occurs in $V^{dir}$ variables due to the action effects. The PDDL keyword \texttt{increase} is overloaded to refer to an encapsulated object~\cite{piacentini2015AI} and the external module is called if the PDDL action to be expanded has an \textit{effect} of the form \texttt{(increase \ $(v^{dir}_i)$ \ $(v^{ind}_j)$)}, where $v^{dir}_i \in V^{dir}$ and $v^{ind}_j \in V^{ind}$. We denote the set of such actions by $A_s$. It is to be noted that the elements of this set can vary depending on the requirements of a particular domain. However, the process for achieving the semantic attachments of the external module remains the same. In this paper, the set $ A_s = \{\texttt{goto\_region}, \texttt{goto\_lift}, \texttt{goto\_room}, \texttt{goto\_door}\} $. Every time a $v^{dir}_i$ is changed due to the direct effects of an action $a_i \in A_s$, the values of the respective $v^{ind}_j$ is calculated by the external module, attaching the computed value to the indirect variable $v^{ind}_j$, thereby updating the state. Once an action $a_i$ is expanded by the task planner, the corresponding start ($s_i$) and goal ($s_{i+1}$) task states are communicated to the motion planner through the the function \texttt{(triggered ?from ?to)} (line 6). For the task state $s_i$, the robot pose $\tau_i(0) =  \phi(s_i)$ is known since it is the mean of the current belief distribution. For the task state $s_{i+1}$, each pose instantiation $s_{i+1}^{n_j} \in \phi(s_{i+1})$ is considered as a goal node (line 9). With $\tau_i(0)$ as the start node, a motion plan is attempted to each of the goal node $s_{i+1}^{n_j}$. The set of feasible motion plans is obtained by performing a search over the roadmap. Along each edge of the roadmap, the belief at $s_i$ is propagated to $s_{i+1}^{n_j}$ by simulating the sequence of controls and observations. We use EKF to compute the appropriate matrices for belief computation as shown in~\ref{eq:predict}. The posterior belief is computed at each node if a landmark is detected by the robot's sensor. \revtext{This belief search process is shown in Algorithm~\ref{algo:bsp}}. The motion costs and the corresponding feasible motion plans are populated to the sets $c$ and $T$ respectively (line 12). The motion plan that corresponds to minimum cost is then computed as $\tau_i^{j^*}$ (lines 14-15). The computed values by the external module is then passed to the respective indirect variables $V^{ind}$ (line 16), achieving semantic attachments. The corresponding motion plan $\tau_i$ and the goal node $s_{i+1}^{n_{j^*}}$ are stored and this goal node subsequently becomes the start node for the roadmap search from $s_{i+1}$. Consequently, the belief estimates returned by the semantic attachments guide the TRPG in identifying the \textit{helpful actions}, besides providing an efficient heuristic evaluation for the task plan.

For the \textit{office domain}, the feasibility of the motion plan $\tau_i^{j^*}$ is checked by accounting for the trace of the covariance matrix upon reaching a cubicle associated with $s_{i+1}$, that is, $trace(\Sigma_{s_{i+1}^{j^*}})$ . Since the cubicle doors are of specific length, we bound the trace by a constant $\eta$. However, the failure of an action $a_i$ to find a feasible motion plan during the current expansion does not mean that it has to be discarded. Feasibility also depends on the sequence of actions performed earlier. A different action sequence prior to $a_i$ can render $a_i$ feasible. Hence infeasible actions are not discarded and are set aside for reattempting later. Consequently the feasibility check is performed for the returned optimal plan $\pi^*$. The plan is feasible if for each $a_i \in \pi^*$, the $trace(\Sigma_{s_{i+1}^{j^*}}) < \eta$; else there is no is feasible plan.
\subsubsection{Cost Function}
\label{sec:cost}

So far we have been agnostic about the cost function used while selecting the nodes for expansion. Though our formulation can be adapted to any generic cost functions we use a standard cost function~\cite{indelman2015IJRR}
\begin{equation}
    c \doteq M_uc_u + M_Gc_G + M_{\Sigma}c_{\Sigma}
\end{equation}
\noindent where $c_u$ is the control usage, $c_{G}$ is the distance to goal and $c_{\Sigma}$ is the cost due to uncertainty, defined as $trace(\Sigma)$, where $\Sigma$ is the state covariance associated with the robot belief. $M_u, M_G$ and $M_{\Sigma}$ are user-defined weights. For the current node $n$ that is considered for expansion, the cost $c$ is computed for each of the nodes that shares an edge with $n$. The node with the minimum $c$ is selected as the next node $n^*$ for expansion. As such, this can be extended to non-myopic planning in a trivial manner, but it is not the current focus of this paper. It is to be noted that $n^*$ is considered only if it is not already in the expanded path with the $n$ being the last node added to the path. So if $n^*$ leads to a cycle, the next best node $n^{**}$ is selected.

As mentioned in the previous \revtext{section}, in case of the \textit{office domain} we add the condition $c_{\Sigma_g} < \eta$, where $_{\Sigma_g}$ is the trace of the goal state covariance and $\eta$ is a constant. The cubicle doors have a width of $2m$ and considering maximum uncertainty along the door width we fix $\eta = 3m^2$ as the maximum upper limit and discard the motion plans with $c_{\Sigma_g} > 3$ (see lines 19-24, Algorithm~\ref{algo:TMP}). For the \textit{corridor domain}, since the automatic doors are of 1$m$ in length, we set an upper bound of $\eta = 0.75m^2$, which corresponds to an uncertainty budget of 0.5$m$ in each of the pose component. This check is performed when the robot is at a node directly in front of the door as a result of executing the action \texttt{goto\_door}. If the estimated covariance is within the uncertainty budget an edge is added between the current node and the nearest node in the next room to which the robot can navigate via the door. Once the robot traverses the door to reach the next room by executing the action \texttt{goto\_room}, the newly added edge is removed from the roadmap. The process of addition and deletion of an edge occur within the external module as a consequence of the \texttt{goto\_door} and  \texttt{goto\_room} actions. 

\subsubsection{Optimality}

For a given roadmap, the plan synthesized by our approach is optimal at the task-level. This means that the task plan cost returned by our approach ($c^*$) is lower than any of the other possible task plan costs ($c$). Let us denote the optimal plan corresponding to $c^*$ as $\pi^*$. Suppose that there exists a plan $\pi$ with associated cost $c$ such that $c < c^*$. If $\pi$ and $\pi^*$ have the same sequence of actions, this is not possible since the action costs are evaluated by the motion planner and for a given roadmap, the motion cost returned is the optimal for each action, giving $c^* \leq c$. If $\pi$ and $\pi^*$ have a different sequence of actions, the task planner ensures that the returned plan is optimal, giving $c^* \leq c$. Therefore, in both the case, we have $c^* \leq c$.

\subsubsection{Completeness}

We provide a sufficient condition under which the probability of our approach returning a plan approaches one exponentially with the number of samples used in the construction of the roadmap. A task planning problem, $\Omega = (S, A, \gamma, s_0, S_g)$ is complete if it does contain any dead-ends~\cite{hoffmann2001JAIR}, that is there are no states from which goal states cannot be reached. The PRM motion planner is probabilistically complete~\cite{karaman2011IJRR}, that is the probability of failure decays to zero exponentially with the number of samples used in the construction of the roadmap. Therefore, if the motion planner terminates each time it is invoked then probability of finding a plan, if it exists, approaches one. 

On the one hand our approach is probabilistically complete; on the other hand, it is also resolution complete since the motion plan feasibility depends on the parameter $\eta$. Nevertheless, given a fixed value of $\eta$, the probability that the planner fails to return a solution, if one exists, tends to zero as the number of samples approaches infinity. In this sense the best that we can guarantee is probabilistic completeness.

%% file: results.tex
In this Section, we validate our approach in two different robot navigation domains, namely \textit{office domain} and \textit{corridor domain} as described in Section~\ref{sec:nav_domain1} and Section~\ref{sec:nav_domain2}. We use the temporal POPF-TIF~\cite{bernardini2017ICAPS} as our task planner by customizing it to achieve semantic attachments of an external module. The external module performs a PRM-based planning in the belief space and is implemented as a dynamically loaded shared library that is passed as an input to the planner. The enumeration into direct variables $V^{dir}$ and indirect variables $V^{ind}$ are listed in the external module. The performance are evaluated on an Intel{\small\textregistered} Core i7-6500U under Ubuntu 16.04 LTS.


First, we present the motion and sensor models used in our experiments\footnote{To simplify the notation, most variables are presented without time indexes.}. Then, we discuss the metrics devised to evaluate the usefulness and validity of our approach. Finally, we present the evaluation of our approach in the two navigation domains using the devised metrics. 

\subsection{Motion and Sensor Model}
The robot dynamics is modeled using the following non-linear model~\cite{thrun2005book} 
\begin{equation}
\begin{split}
x_{k+1}(1) & = x_{k}(1) + \delta_{trans} \cdot \cos(x_{k}(3)+ \delta_{rot1})\\
x_{k+1}(2) & = x_{k}(2) + \delta_{trans} \cdot \sin(x_{k}(3)+ \delta_{rot1})\\
x_{k+1}(3) & = x_{k}(3) + \delta_{rot1}+ \delta_{rot2}
\end{split}
\label{odometry_model}
\end{equation}

\noindent where $x_k\doteq(x, y, \theta)$, is the robot pose at time $k$ with $x_{k}(1) = x, x_{k}(2) = y$ and $x_{k}(3) = \theta$ and $u_k \doteq (\delta_{rot1}, \delta_{trans}, \delta_{rot2})$ is the applied control. The model in (\ref{odometry_model}) assumes that the robot ideally implements the following commands in order: rotation by an angle of $\delta_{rot1}$, translation of $\delta_{trans}$ and a final rotation of $\delta_{rot2}$ orienting the robot in the required direction\footnote{The state transition model form of (\ref{odometry_model}) is given in (\ref{eq:odometry_model}).}. It is to be noted that the robot accrue translational and rotational errors while executing $u_k$.

In the EKF, the Jacobian of the state transition model with respect to the state $x_k$ denoted by $F_k$ (see (\ref{eq:predict}) and (\ref{eq:update})) is obtained by linearizing the state transition function about the mean state at $x_k$ and is given by

\begin{equation}
F_k= \begin{bmatrix}
\frac{\partial f}{\partial x_k(1)} & \frac{\partial f}{\partial x_k(3)}  & \frac{\partial f}{\partial x_k(3)}  \\ 
\frac{\partial f}{\partial x_k(1)} & \frac{\partial f}{\partial x_k(3)}  & \frac{\partial f}{\partial x_k(3)}  \\
\frac{\partial f}{\partial x_k(1)} & \frac{\partial f}{\partial x_k(3)}  & \frac{\partial f}{\partial x_k(3)}  \\
\end{bmatrix}
= \begin{bmatrix}
1 & 0  & -\delta _{trans} \cdot \sin(x_{k}(3)+ \delta_{rot1})  \\ 
0 & 1 & \delta _{trans} \cdot \cos(x_{k}(3)+ \delta_{rot1}) \\
0 & 0  & 1  \\
\end{bmatrix}
\label{F_matrix}
\end{equation}
Similarly, the linearized process noise, $R_k = V_kW_kV_k^T$, is obtained by computing the Jacobian of $V_k$

\begin{equation}
\resizebox{.9\hsize}{!}{$
V_k= \begin{bmatrix}
\frac{\partial f}{\partial \delta_{rot1}} & \frac{\partial f}{\partial \delta_{trans}}  & \frac{\partial f}{\partial \delta_{rot2}} \\ 
\frac{\partial f}{\partial \delta_{rot1}} & \frac{\partial f}{\partial \delta_{trans}}  & \frac{\partial f}{\partial \delta_{rot2}}  \\
\frac{\partial f}{\partial \delta_{rot1}} & \frac{\partial f}{\partial \delta_{trans}}  & \frac{\partial f}{\partial \delta_{rot2}}  \\
\end{bmatrix}
= \begin{bmatrix}
-\delta_{trans} \cdot \sin(x_{k}(3) + \delta_{rot1}) & \cos(x_{k}(3) + \delta_{rot1}) & 0 \\ 
\delta_{trans} \cdot \cos(x_{k}(3) + \delta_{rot1}) &  \sin(x_{k}(3) + \delta_{rot1})  & 0   \\ 
1   & 0  & 1 \\
\end{bmatrix}
$}
\end{equation}

The noise covariance matrix $W_k$ is formulated as below with $\alpha_1$ to $\alpha_4$ being the robot-specific error parameters~\cite{thrun2005book} modeling the accuracy of the robot motion

\begin{equation}
\resizebox{.87\hsize}{!}{$
W_k= 
\begin{bmatrix}
\alpha_1 \cdot \delta_{rot1}^2 + \alpha_2 \cdot \delta_{trans}^2 & 0  &0  \\ 
0 & \alpha_3 \cdot \delta_{trans}^2 + \alpha_4 \cdot (\delta_{rot1}^2 +\delta_{rot2}^2)  & 0   \\ 
0   & 0  & \alpha_2 \cdot \delta_{trans}^2 + \alpha_1 \cdot \delta_{rot2}^2 \\
\end{bmatrix}
$}
\end{equation}

As for the sensor model, we use a landmark-base model

\begin{equation}
z_k  = \begin{bmatrix} 
r    =  \sqrt{(l_i(1)-x_k(1))^2+(l_i(2)-x_k(2))^2}\\ \\
\phi  =  \arctan(\frac{l_i(2)-x_k(2)}{l_i(1)-x_k(1)})  - x_{k}(3) \  
\end{bmatrix} + v_k \
,  \ v_k \sim \mathcal{N}(0,Q_k)
\label{sensor_model}
\end{equation}
\noindent where $r$ and $\phi$ are the range and bearing of the $i$-th landmark $l_i$ relative to the robot frame. The sensor model is linearized to obtain the Jacobian $H_k$, which is the partial derivative of the measurement function with respect to the robot state\footnote{The measurement function form of (\ref{sensor_model}) is given in (\ref{eq:measurement_model}).}.
\begin{equation}
H_k= \begin{bmatrix}
\frac{\partial r}{\partial x_k(1)} & \frac{\partial r}{\partial x_k(2)}  & \frac{\partial r}{\partial x_k(3)}  \\ 
\frac{\partial \phi}{\partial x_k(1)} & \frac{\partial \phi}{\partial x_k(2)}  & \frac{\partial \phi}{\partial x_k(3)}  
\end{bmatrix}
= \begin{bmatrix}
-\frac{(l_i(1)- x_k(1))}{r} & -\frac{(l_i(2)- x_k(2))}{r}  &0  \\ 
\frac{ (l_i(2)- x_k(2)) }{r^2} & -\frac{(l_i(1)- x_k(1))}{r^2}  & -1  \\ 
\end{bmatrix}
\end{equation}

We would like to reiterate the fact that since we are in the planning phase, the nominal observation $\hat{z} = h(x, l_i)$ is corrupted with noise to simulate future observations.  
\subsection{Plan Metrics}
To benchmark our approach we consider four different \revtext{cost formulations} that differ in their motion cost computation \revtext{and thereby the task-level action costs}. Though our formulation can be adapted to any general cost function (see Section~\ref{sec:cost}), we choose the following four \revtext{cost functions} to demonstrate the efficiency of our approach:

\begin{itemize}

\item \textit{\revtext{Euclidean cost}}: The motion planner is never called and the task cost are evaluated computing the Euclidean distance $c_{euc}$ between the geometric instantiations of $s_i$ and $s_{i+1}$, that is, between $\tau_i^j(0)$ and $\tau_i^j(1)$. Here $ c \doteq c_{euc}$.

\item \textit{\revtext{$\sigma-$Euclidean cost}}: This configuration evaluates the motion cost as the sum of Euclidean distance between $\tau_i(0)$ and $\tau_i(1)$ and the cost due to uncertainty, defined as $c_{\Sigma}=trace(\Sigma)$, where $\Sigma$ is the covariance at each node of $\tau_i$. The general form of this cost function is $c \doteq M_{euc}c_{euc} + M_{\Sigma}c_{\Sigma} $.

\item \textit{\revtext{PETLON cost}}: In this configuration, the motion planner returns the trajectory length or the geometric-level cost of traversing from $s_i$ to $s_{i+1}$, that is, from $\tau_i^j(0) \in \phi(s_i)$ to $\tau_i^j(1) \in \phi(s_{i+1})$. The general form of the cost for this configuration is $c \doteq M_uc_u + M_Gc_G $, where $c_u$ is the control usage and $c_{G}$ is the distance to goal. Since we assume straight line path between two sampled poses, the applied control for translation, that is $\delta_{trans}$ represents the trajectory length. \revtext{We note here that the motion planner in PETLON~\cite{lo2018AAMAS} computes the geometric-level cost of traversing from one state to another and hence this configuration will be used to compare MPTP with PETLON}.

\item \textit{\revtext{MPTP cost}}: In this configuration, we use the cost function as defined in Section~\ref{sec:cost}, that is, $c \doteq M_uc_u + M_Gc_G + M_{\Sigma}c_{\Sigma}$, where $c_u$ is the control usage, $c_{G}$ is the distance to goal and $c_{\Sigma}$ is the cost due to uncertainty. \revtext{It is noteworthy that \textit{PETLON cost} is subsumed in \textit{MPTP cost} since \textit{MPTP cost} is fundamentally \textit{PETLON cost} added with the cost due to uncertainty}. 
\end{itemize}


\subsection{Office Domain}
\begin{figure}[h]
	\centering
		\includegraphics[scale=0.29]{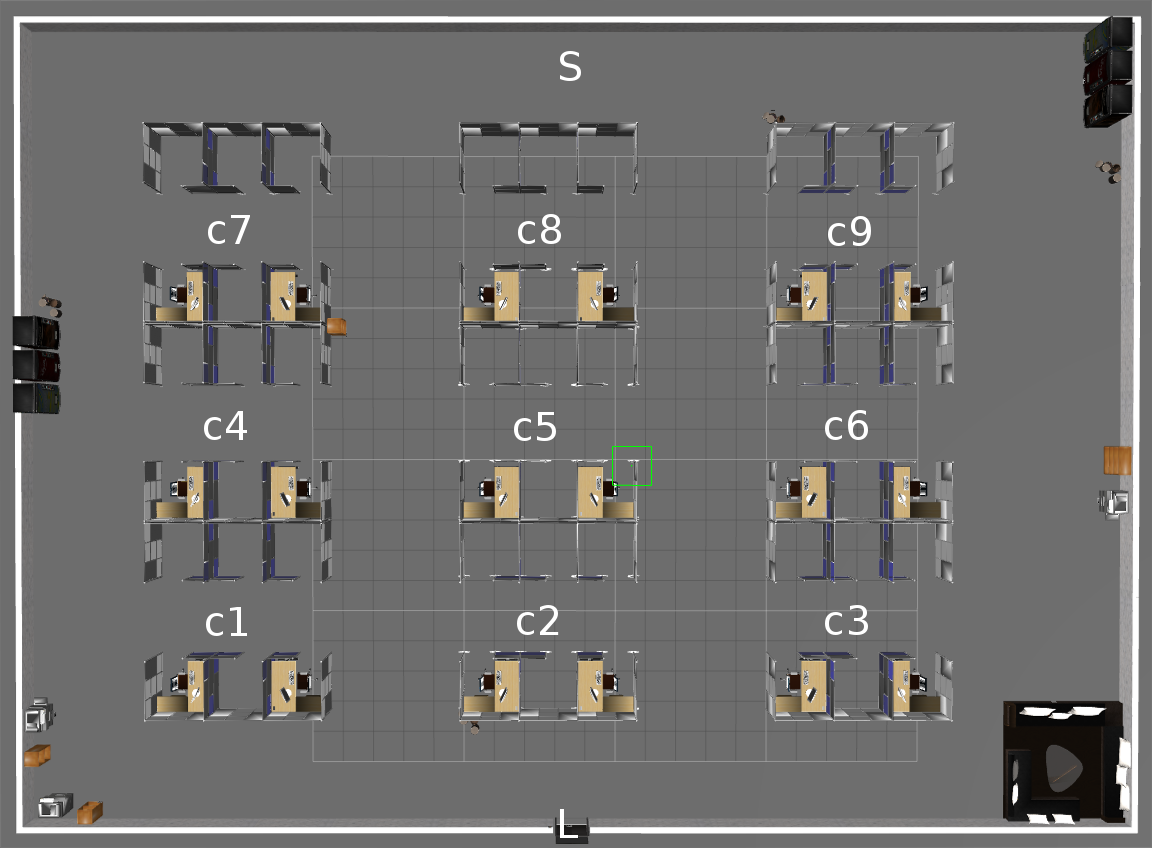}
		\caption{\revtext{Top view of the simulated environment in Gazebo. See \textit{office domain} in Section~\ref{sec:approach} for a detailed description.}}
	\label{fig:gazebo}
\end{figure}

This domain is simulated in Gazebo~\cite{koenig2004IROS} by constructing an office environment of $36m \times 25m$; top view of the simulated environment is shown in \revtext{Fig.~\ref{fig:gazebo}. We note here that the landmarks considered in this domain are the objects outside the cubicles like printers, trash cans, lounge, vending machines and book-shelves}. The robot is required to collect documents from different cubicles, and the documents are then taken to the next floor via the lift $L$. 

\subsubsection{Validation}
\label{subsec:cost}

We first demonstrate the need for a combined TMP for navigation. Unless otherwise stated, the panning times presented is an average for 25 different planning sessions. Consider the following scenario in which the robot is required to collect documents from the cubicles $c3$, $c4$, $c6$ and $c9$. We first run the planner with \textit{\revtext{Euclidean cost}} to synthesize the task plan. We remind that in this configuration the motion planner is never called and the action costs are evaluated by considering the Euclidean distance between the start and goal regions. The plan synthesized is $S\rightarrow c3 \rightarrow c4 \rightarrow c6 \rightarrow c9 \rightarrow L$.
This plan is then given to the motion planner, to compute the corresponding cost due to uncertainty \revtext{$c_{\Sigma}$ which is the trace of the robot state covariance}. The task planning cost and the motion planning cost are added to estimate the overall planning cost, which equated to 298.84. \revtext{The addition of the two costs is possible because we first compute the task plan which is then passed to the motion planner to compute the cost due to uncertainty. Therefore the overall planning cost is the task planning cost combined with motion planning cost}. In the same way, the overall planning time was computed to be 0.94 \revtext{($\pm 0.09$)} seconds \revtext{by adding the time for task planning and motion planning, respectively}. Next, we ran the planner with \textit{\revtext{$\sigma-$Euclidean cost}}, returning the plan $S\rightarrow c4 \rightarrow c9 \rightarrow c6 \rightarrow c3 \rightarrow L$, in 1.28 \revtext{($\pm 0.06$)} seconds with a total cost of 90.89. This configuration evaluates the motion cost as the sum of Euclidean distance and the cost due to uncertainty. It is seen that there is a significant difference in the plan quality as the cost is improved by a factor of 3 for \textit{\revtext{$\sigma-$Euclidean cost}}. This difference in cost is attributed to the different task sequence synthesized. \revtext{Essentially, \textit{Euclidean cost} corresponds to planners that pre-compute motion costs of all task-level actions or use an admissible heuristic for the same (for example, the approach in~\cite{wong2018optimal}). The task plan is then given to the motion planner for execution, assuming that such a motion plan exists. In contrast, \textit{\revtext{$\sigma-$Euclidean cost}} checks for the motion feasibility and estimates the motion costs while expanding each task-level action and thus corresponds to an integrated TMP approach as discussed in this paper.} \revtext{The difference in plan quality between \textit{\revtext{Euclidean cost}} and \textit{\revtext{$\sigma-$Euclidean cost}}} clearly \revtext{demonstrates} the efficiency of a combined TMP approach as opposed to performing \revtext{task planning and motion planning} separately. Though our considered scenario is much less knowledge-intensive than real-world scenarios, \revtext{the above example conveys the need for a combined task-motion planner.}

\begin{figure}[h!]
	\centering
	
\subfloat{\includegraphics[scale=0.146]{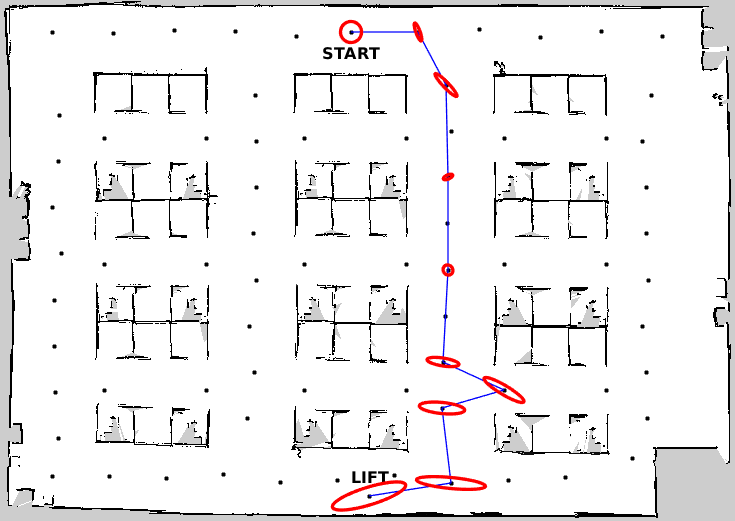}\label{fig:1}}
	\hspace{0.02cm}
\subfloat{\includegraphics[scale=0.147]{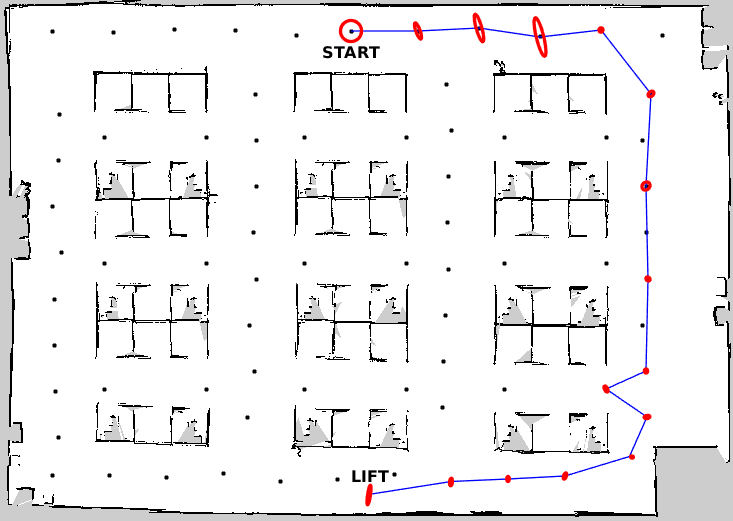}\label{fig:2}}
\hspace{0.02cm}
\subfloat{\includegraphics[scale=0.16]{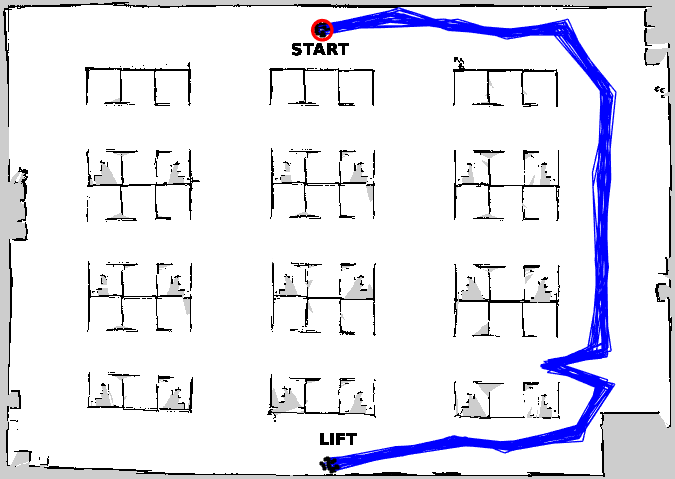}\label{fig:3}}
\vspace{0.05cm}
		\caption{(\textit{left} and \textit{center}) The propagated belief distributions along the planned paths for \textit{\revtext{PETLON cost}} and \textit{\revtext{MPTP cost}}. The belief estimates for a single planning instantiation corresponding to a unique set of simulated observations are shown. Black dots represent the sampled poses. (\textit{left}) Shortest path route that corresponds to \textit{\revtext{PETLON cost}}. (\textit{center}) Belief space planning corresponding to \textit{\revtext{MPTP cost}}, returning an information rich route. (\textit{right}) Traces of robot's true state while starting from the initial belief-- run \revtext{with} \textit{\revtext{MPTP cost}}.}
		\label{fig:BSP}
\end{figure}

Next, we run the planner with \textit{\revtext{PETLON cost}} and \textit{\revtext{MPTP cost}} to demonstrate the advantage of planning in belief space, that is using \revtext{our} MPTP approach. \revtext{We recall here that similar} to PETLON~\cite{lo2018AAMAS}, \revtext{with} \textit{\revtext{PETLON cost}}, the motion planner evaluates the geometric-level cost of traversing $\tau_i(0)$ to $\tau_i(1)$, whereas  \revtext{with} \textit{\revtext{MPTP cost}}, in addition to considering the geometric-level cost of traversing, the cost due to uncertainty is \revtext{also incorporated}. We consider a scenario in which the robot has to collect a document from cubicle $c3$. The planned trajectories in both the scenarios with the corresponding covariance estimated at each node (only the ($x$,$y$) portion is shown) is shown in Fig.~\ref{fig:BSP}. Clearly, the belief space \revtext{task-motion} planner (\textit{\revtext{MPTP cost}}) returns a route which is rich in sensor information (see Fig.~\ref{fig:BSP} in the mid), enabling effective localization. \revtext{\textit{\revtext{PETLON cost}} returns the shortest path trajectory but with an increased robot state uncertainty}. Fig.~\ref{fig:BSP} on the right hand side shows the  traces of true robot state for 25 different simulations while running on \textit{\revtext{MPTP cost}}---the initial state being sampled from the known initial belief. 

\setlength{\tabcolsep}{4pt}
	\begin{table}
		{ 
			\revtext{\begin{tabular}{l l  ccc | ccc }                             
			\hline 	
				\multicolumn{1}{c}{\multirow{ 1}{*}{}} &  \multicolumn{1}{c}{$d$} & \multicolumn{3}{c}{Overall time (s)} & \multicolumn{3}{c}{Cost}  \\
								{} & {} & \multicolumn{1}{c}{$\texttt{c}=2$} &  \multicolumn{1}{c}{$\texttt{c}=4$} & \multicolumn{1}{c}{$\texttt{c}=6$} & \multicolumn{1}{c}{$\texttt{c}=2$} &  \multicolumn{1}{c}{$\texttt{c}=4$} & \multicolumn{1}{c}{$\texttt{c}=6$} \\
				
				\hline
				\textit{MPTP cost}	 &1 & 1.34 $\pm$ 0.05 & 2.24 $\pm$ 0.15& - & 83.84 & 90.27 & -\\
                     & 1.5 & 3.41 $\pm$ 0.08 & 7.16 $\pm$ 0.12 & 14.04 $\pm$ 0.09 & 88.18 & 101.01 & 237.59\\
                     & 2 & 9.11 $\pm$ 1.17 & 28.48 $\pm$ 1.19& 46.15 $\pm$ 2.23 & 92.32 & 126.96 & 260.092\\         

                     \hline		
                     
                     {} & {} & \multicolumn{1}{c}{$\texttt{c}=2$} &  \multicolumn{1}{c}{$\texttt{c}=4$} & \multicolumn{1}{c}{$\texttt{c}=6$} & \multicolumn{1}{c}{$\texttt{c}=2$} &  \multicolumn{1}{c}{$\texttt{c}=4$} & \multicolumn{1}{c}{$\texttt{c}=6$} \\	
		       \hline              
\textit{PETLON cost}	 & 1 & 0.47 $\pm$ 0.02 & 0.77 $\pm$ 0.04 & 1.77 $\pm$ 0.01 & 47.80 & 84.88 & 161.47\\
                     & 1.5 & 3.17 $\pm$ 0.03 & 4.91 $\pm$ 0.02 & 7.10 $\pm$ 0.10 & 55.77 & 95.74 & 174.90\\
                     & 2 & 6.08 $\pm$ 0.11 & 9.86 $\pm$ 0.17& 15.14 $\pm$ 1.09& 56.19 & 95.77 & 181.06\\                  
                     \hline
		\end{tabular}}}                                          
		\caption{Overall planning time and cost returned \revtext{while running the task-motion planner with \textit{\revtext{MPTP cost}} and \textit{\revtext{PETLON cost}}}. \revtext{The average number of samples per square meter is denoted by $d$}. $\texttt{c}$ = 2, 4 and 6 denotes the number of cubicles to be visited, increasing the task-level complexity. \revtext{'-' denotes the fact that no plan is found as the condition $\eta < 1$ is violated}.} 
\label{table:result1}
	\end{table}

\subsubsection{Scalability}
\label{subsec:scalability}

\begin{figure}[]
	\centering
		\includegraphics[scale=0.5]{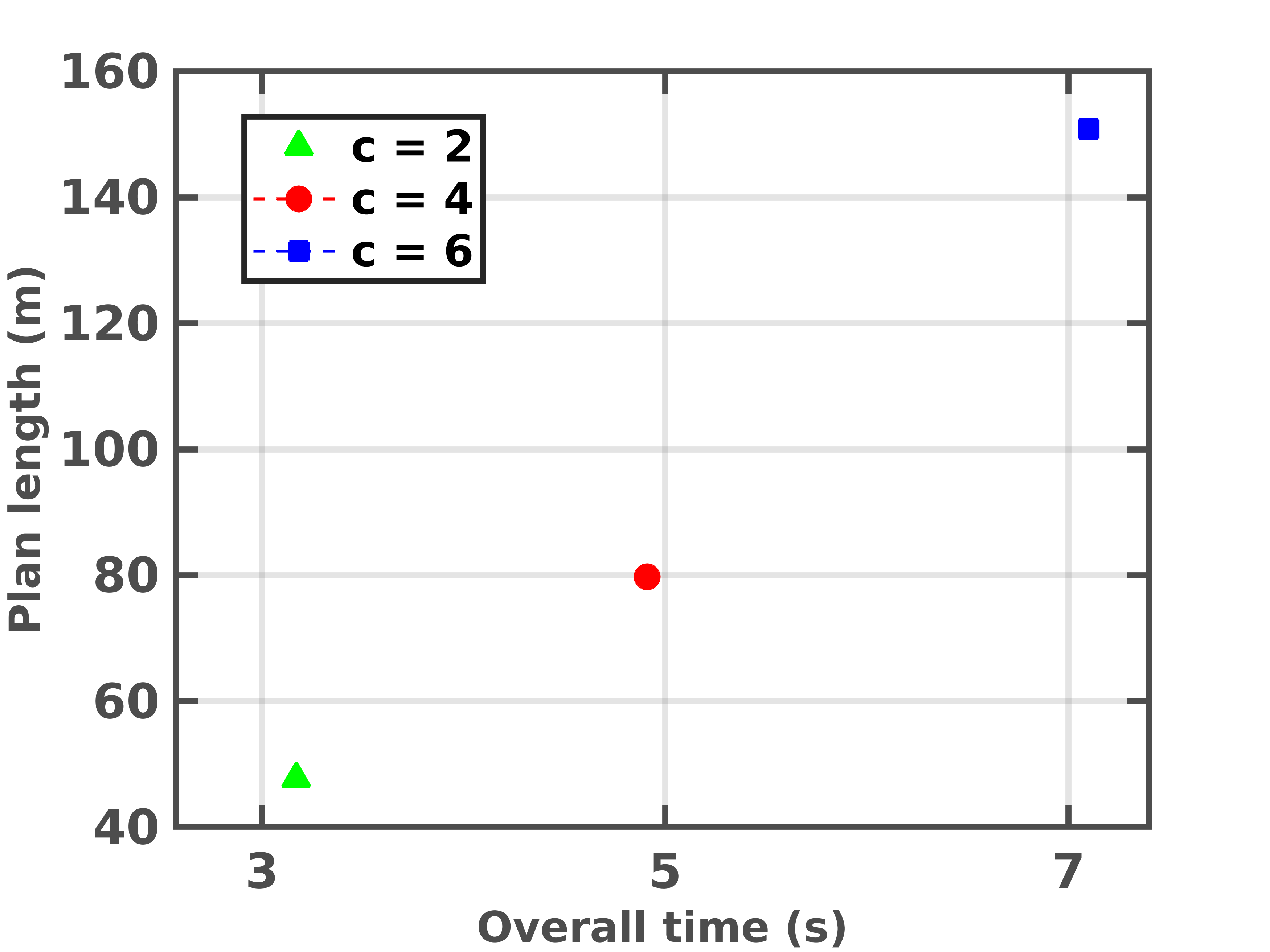}
		\caption{\revtext{Plan length with overall planning time. MPTP is run with \textit{PETLON cost} and a sampling density of $d=1.5$}.}
	\label{fig:scale}
\end{figure}

We test the scalability of our approach by increasing the task-level complexity. We run our planner on three different scenarios where 2, 4, 6 number of cubicles ($\texttt{c}=2,4,6$) are to be visited to collect the corresponding number of documents. This results in evaluating more task-level actions, escalating the task level complexity. We also test these scenarios on varying levels of sample densities. We choose $d=1,1.5,2$, where $d=i$ corresponds to an average of $i$ samples per square meter. The tests are run using \textit{\revtext{MPTP cost}} and \textit{\revtext{PETLON cost}}. The overall planning time and the returned cost can be seen in Table~\ref{table:result1}. \revtext{While we ran with the \textit{MPTP cost}}, for $d = 1$ and $\texttt{c}=6$, \revtext{no feasible motion plan is found since} the condition $\eta < 1$ is violated. However, for higher sample densities, a feasible motion plan is found. The plan quality is increased with increase in $d$, but at the expense of exponentially increasing computation time. It is clearly seen that for our considered scenario $d=1.5$ can be chosen, without much loss of plan quality.

\begin{figure}[t]
  \subfloat[]{\includegraphics[scale=0.30]{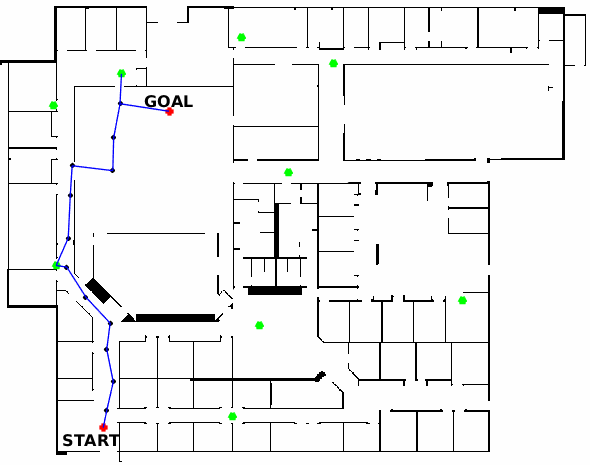}}
  \subfloat[]{\includegraphics[scale=0.46]{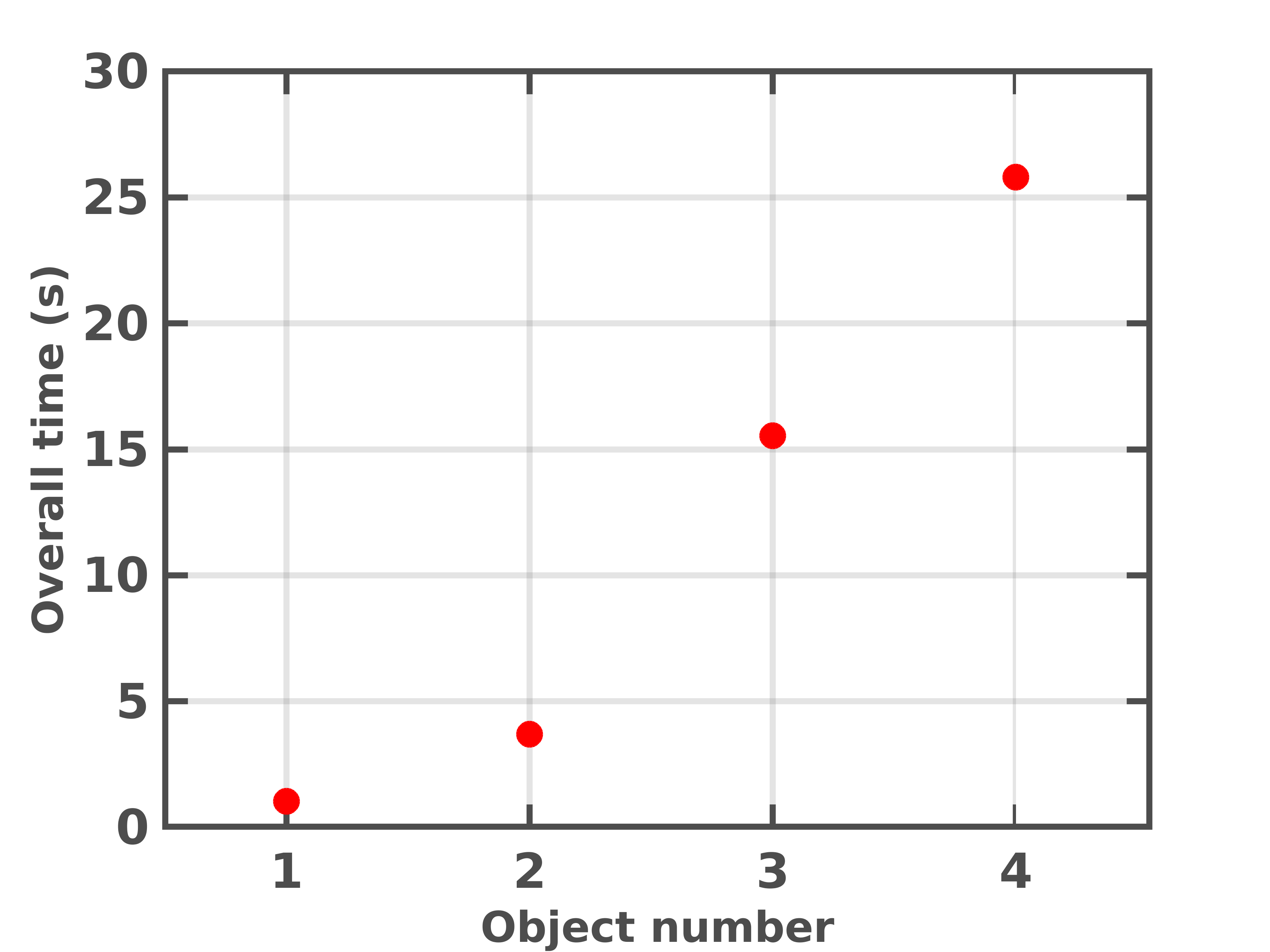}}
  \caption{(a) \revtext{Willow Garage world with nine objects whose instances are marked as green blobs. The optimal path when two objects are to be collected is shown in blue. The planner is run with \textit{PETLON cost}}. (b) Overall planning time with increasing number of objects to be collected for delivering.}
  \label{fig:willow}
\end{figure}

In ~\cite{lo2018AAMAS}, TMP for navigation is performed by evaluating the geometric cost of traversing. \revtext{They} consider a scenario in which \revtext{nine objects are placed at different locations.} \revtext{Two objects from among them are to be collected and delivered to a person such that the geometric cost of traversing is minimum. They report a total planning time of about 15 seconds with a plan length of 37 m. Though the environment considered in~\cite{lo2018AAMAS} is larger than ours, to provide a comparison with PETLON, we run our task-motion planner with \textit{PETLON cost} and evaluate the planning time with respect to the plan length}. In comparison, MPTP \revtext{with \textit{PETLON cost}} fares superiorly with respect to increased task-level complexity. \revtext{To demonstrate this, we first consider three scenarios where 2, 4, and, 6 documents are to be collected to be delivered to the next floor. The results can be seen in Table~\ref{table:result1} under the \textit{PETLON cost} section. We note here that for $d=1.5$ and collecting 6 documents ($\texttt{c}=6$) MPTP \revtext{with \textit{PETLON cost}} took only about 7 ($\pm 0.34$) seconds with a plan length of about 150 m (see Fig.~\ref{fig:scale})}. To provide a better comparison, we also evaluate our approach by considering a much larger environment, the Willow Garage world of $58m \times 45m$ as shown in Fig.~\ref{fig:willow}(a). In this example, \revtext{the robot (at start) needs to collect any two objects from among nine different objects (location of objects marked as green blobs), and deliver it} to a person at the goal location (shown in red). We ran our planner with \textit{\revtext{PETLON cost}}, returning an optimal plan of length 53.94 m in 3.69 \revtext{($\pm$ 0.09)} seconds. \revtext{We recall here that for the same scenario, PETLON~\cite{lo2018AAMAS} report a planning time of about 15 seconds for a plan length of 37 m. In contrast, MPTP with \textit{PETLON cost} is almost three faster. This clearly elucidates the superiority of our approach. PETLON first computes a task plan using an admissible heuristic which is then sent to the motion planner for actual cost evaluation. This cost refinement process is iterated until the optimal plan is found.  MPTP does not require such an iteration since it evaluates the motion cost using semantic attachments as the action is expanded by the task planner}.  The scalability to increasing task complexity is tested by varying the number of objects to be collected (see Fig.~\ref{fig:willow}(b)). The task in which four objects are to be collected was completed in only about 25 \revtext{($\pm$ 1.64)} seconds. \revtext{Therefore MPTP reveals to be much faster than PETLON and is robust to the increasing number of objects and map size.}

\begin{figure}[]
	\centering
		\includegraphics[scale=0.5]{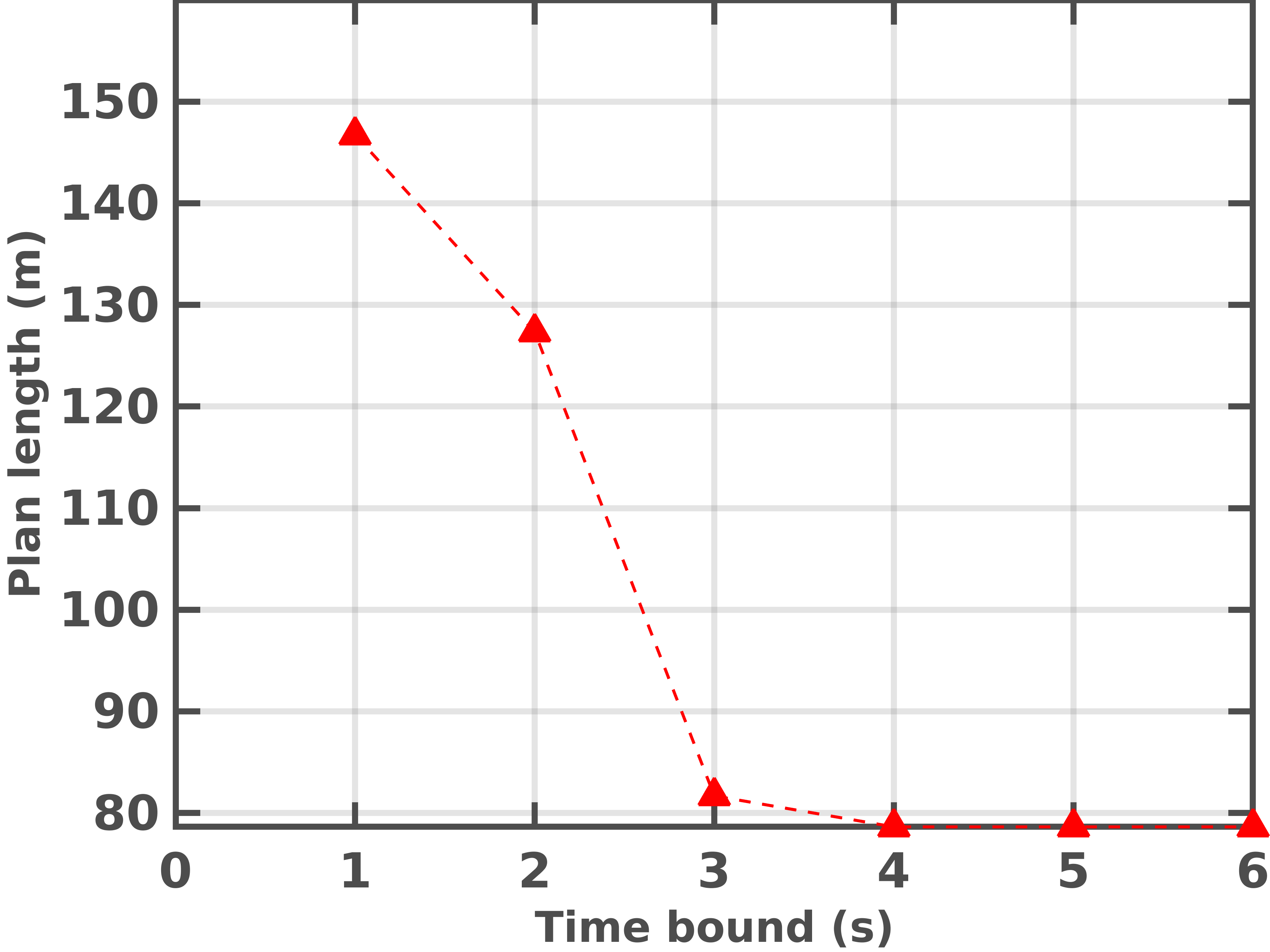}
		\caption{\revtext{Anytime property of MPTP. Valid solutions are returned even when strict bounds are placed on the planning time.}}
	\label{fig:anytime}
\end{figure}

\revtext{POPF-TIF supports \textit{anytime} planning which means that the planner searches for improved solutions until it has exhausted the search space or is interrupted. Specifically, POPF-TIF is run with a \texttt{-n} flag to activate anytime search. A time bound may be specified with the flag \texttt{-tx}, where \texttt{x} is the time bound in seconds and is used in situations with strict time bounds where optimality is sacrificed. We demonstrate this by considering the Willow Garage world in which the robot needs to collect any three documents from among the nine objects and deliver it to a person. We start with a time bound of 1 second and increment it by a second until an optimal solution is found. The result is plotted in Fig.~\ref{fig:anytime}. As the time bound is incremented, the plan quality is increased and for a planning time bound of 4 seconds, the optimal plan length of 78.63 m is returned.}

\begin{figure}[h]
	\centering
		\includegraphics[scale=0.2]{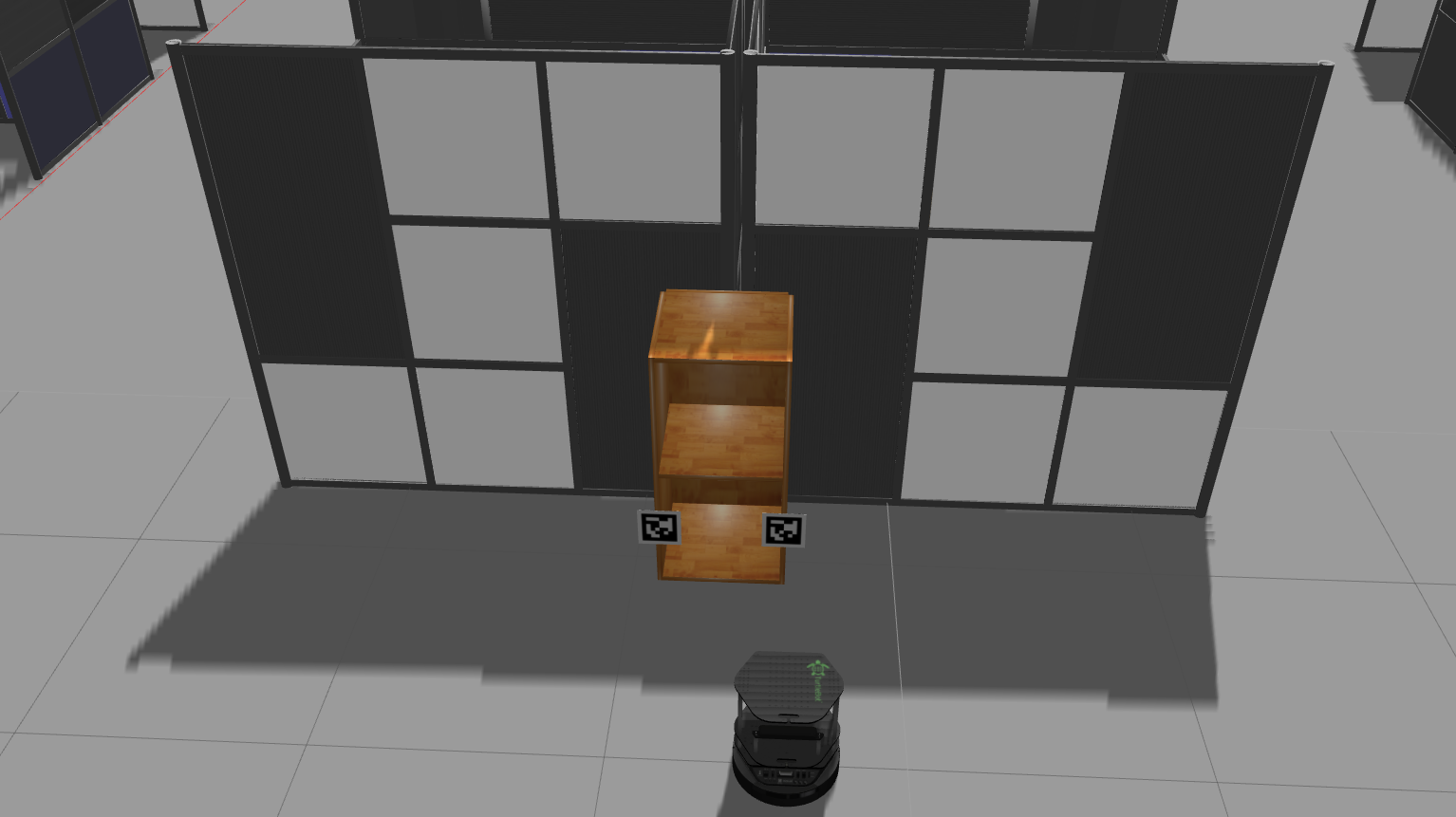}
		\caption{\revtext{A robot in front of AprilTags which provide the transformation between the robot pose and the landmark pose.}}
	\label{fig:april}
\end{figure}

\begin{figure}[]
  \subfloat[]{\includegraphics[trim=0.5cm 0cm 0.5cm 0cm, clip=true,scale=0.19]{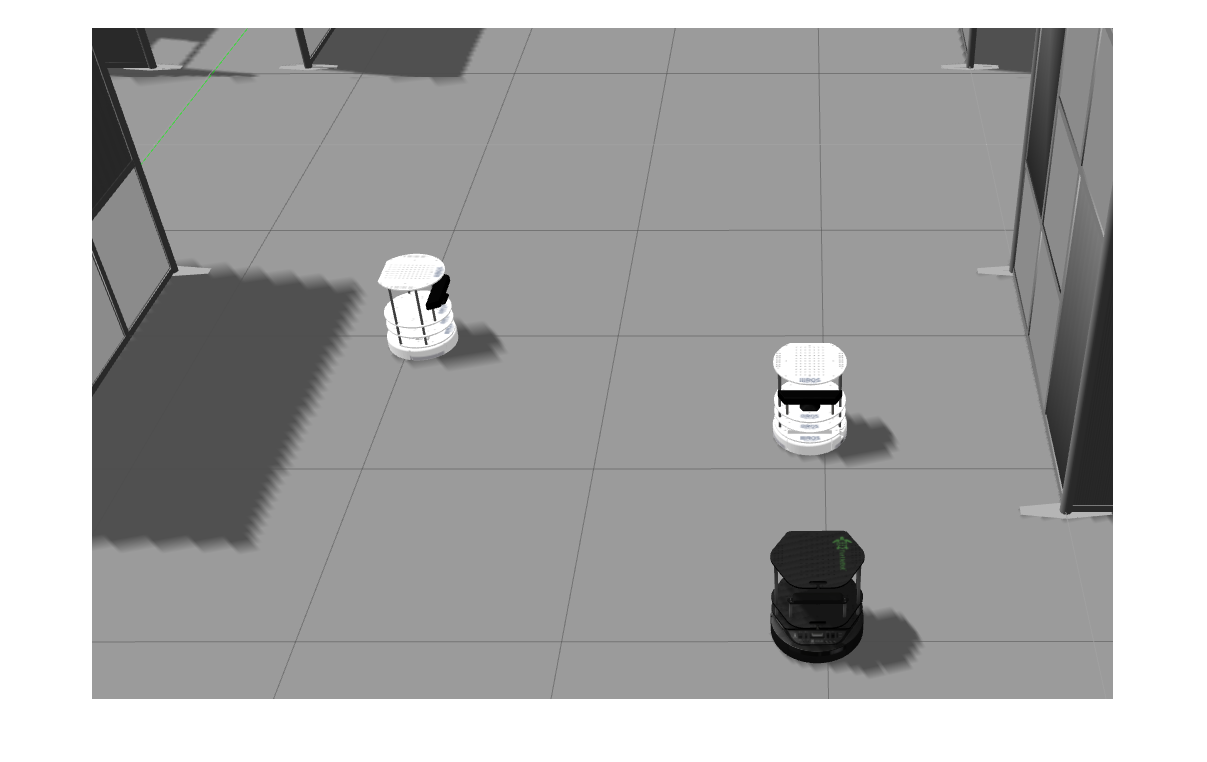}}\hfill
  \subfloat[]{\includegraphics[trim=0.5cm 0cm 0.5cm 0cm, clip=true,scale=0.19]{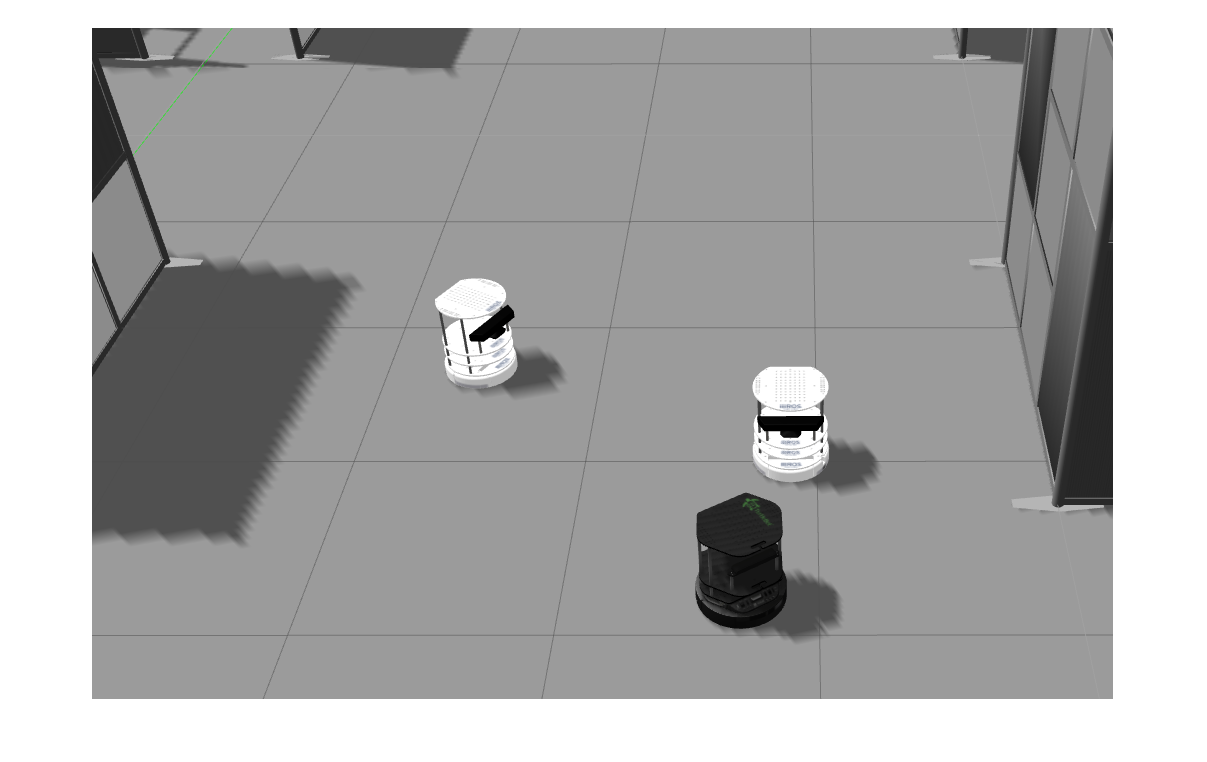}}\\
    \subfloat[]{\includegraphics[trim=0.5cm 0cm 0.5cm 0cm, clip=true,scale=0.19]{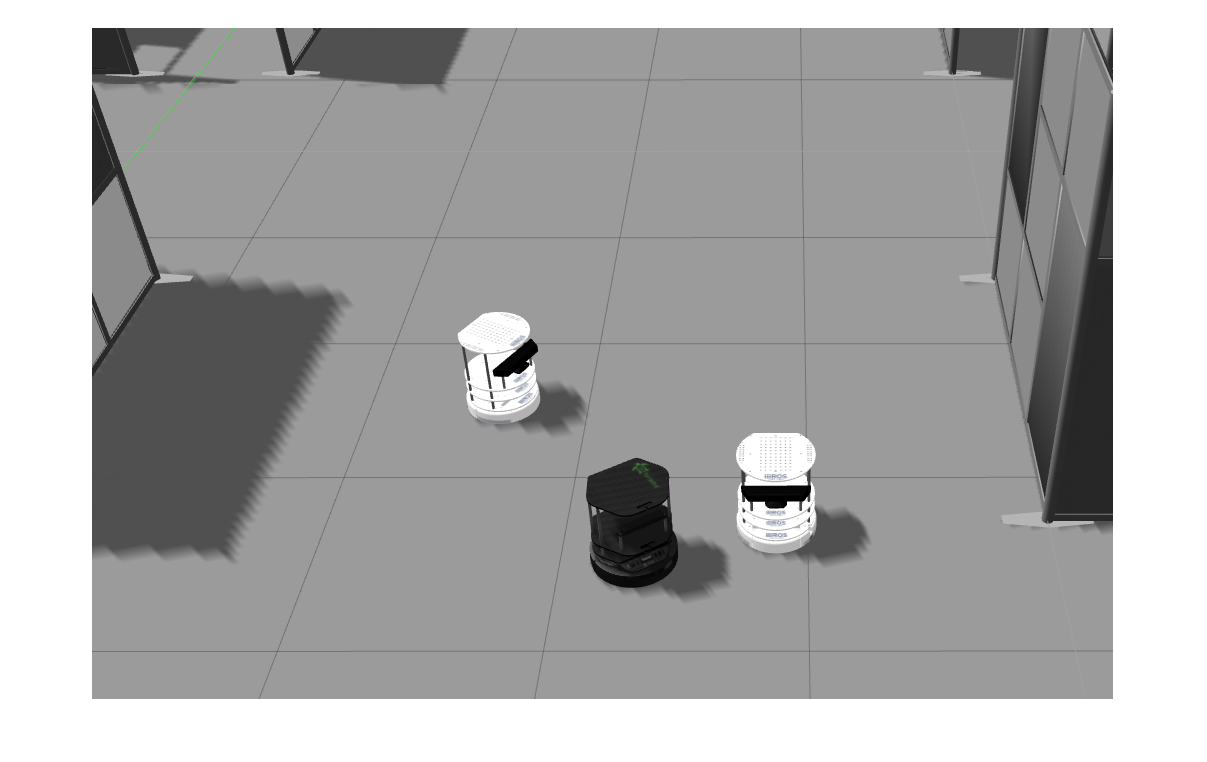}}\hfill
      \subfloat[]{\includegraphics[trim=0.5cm 0cm 0.5cm 0cm, clip=true,scale=0.19]{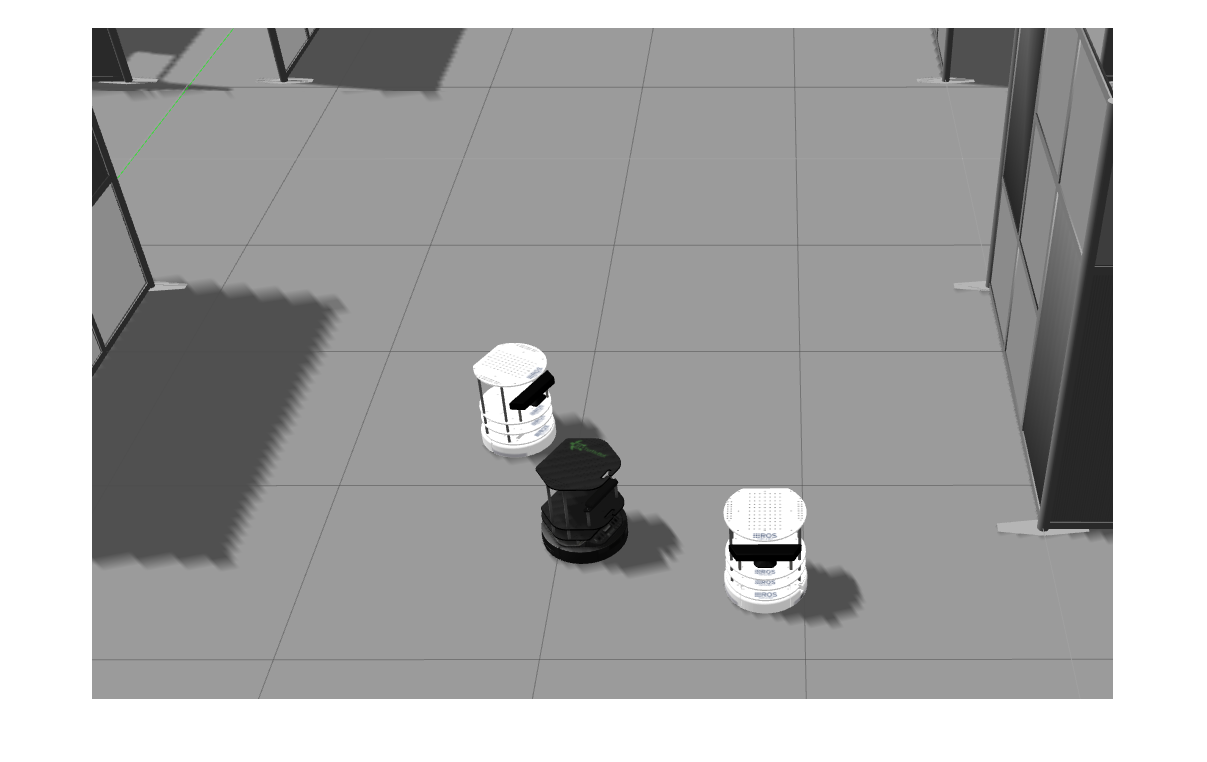}}\\
        \subfloat[]{\includegraphics[trim=0.5cm 0cm 0.5cm 0cm, clip=true,scale=0.19]{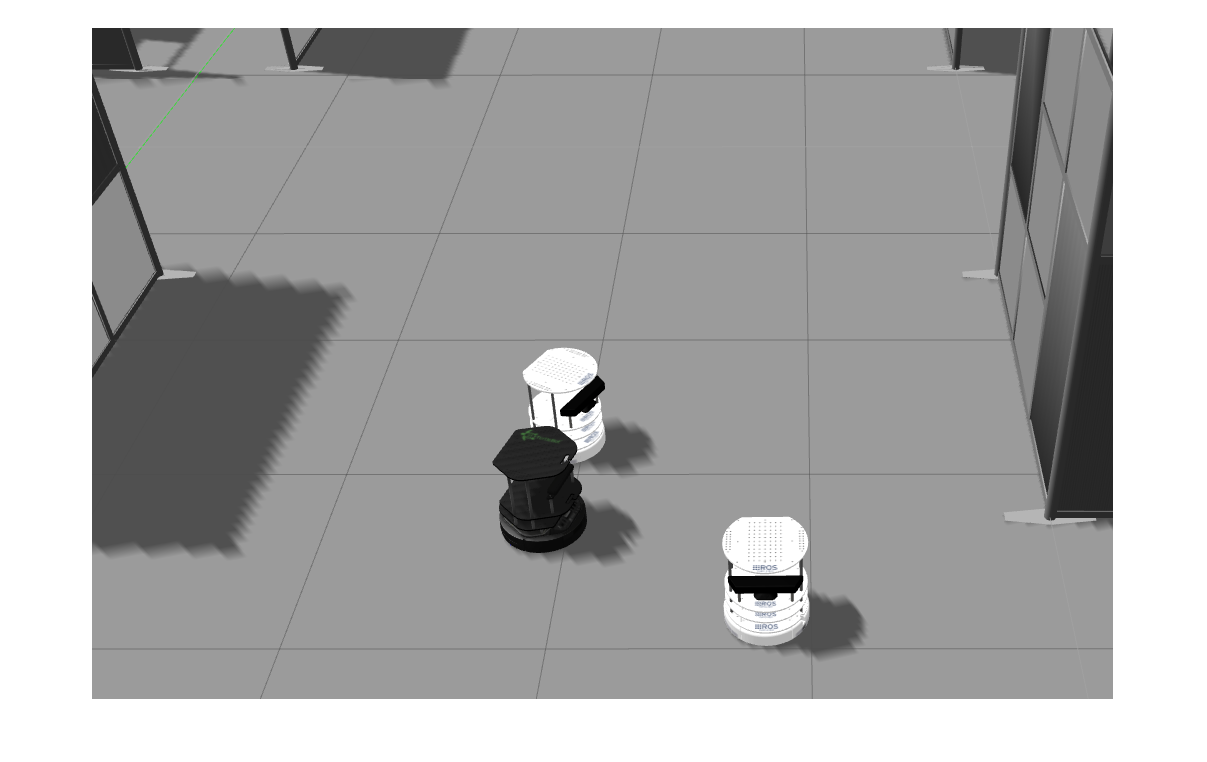}}\hfill
          \subfloat[]{\includegraphics[trim=0.5cm 0cm 0.5cm 0cm, clip=true,scale=0.19]{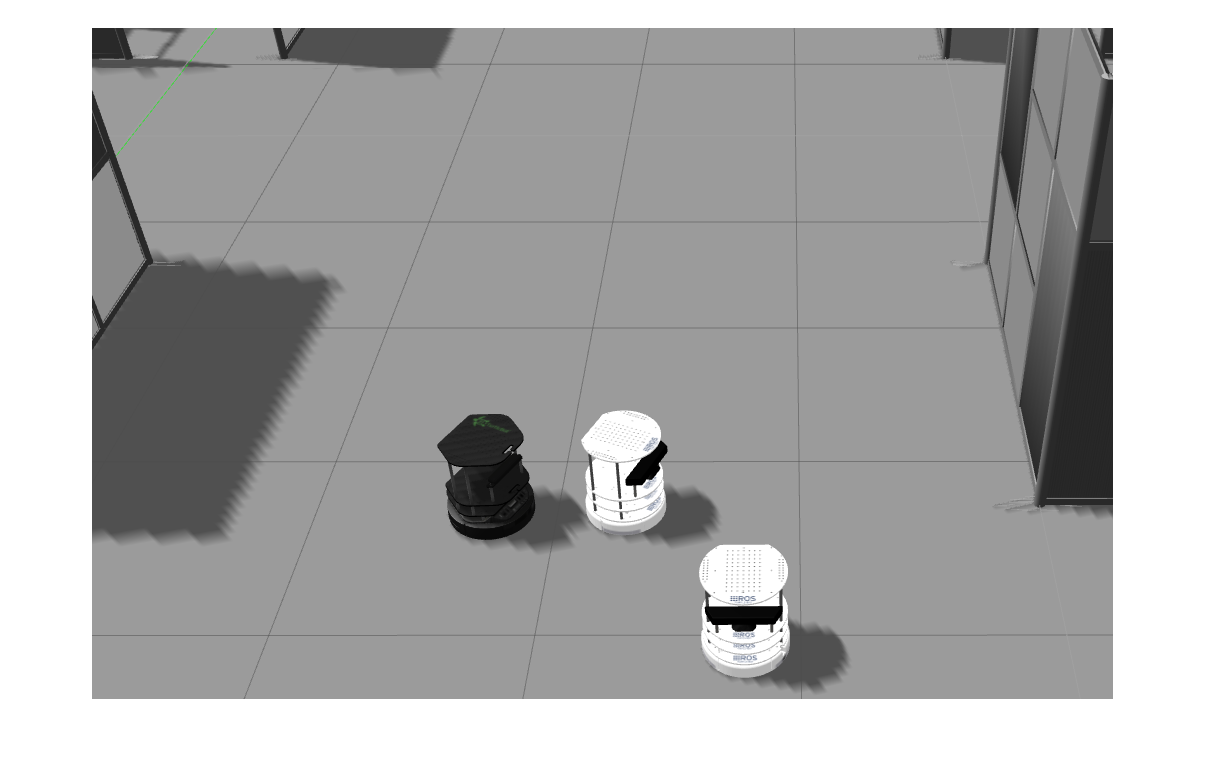}}
  \caption{\revtext{A robot avoiding a couple of dynamic obstacles (white TurtleBot robots) during execution. Our approach is not restrictive to any particular execution strategy and any approach that employs dynamic obstacle avoidance may be used.}}
  \label{fig:execution}
\end{figure}

\revtext{We stress here the fact that in this paper we are mainly concerned with planning and the synthesized plans are given to the robot for execution. Thus, any such execution approach may be employed. In this work, the} generated plans are executed with a TurtleBot robot in the simulated Gazebo environment. We use AprilTags~\cite{olson2011ICRA} to identify objects like printers, trash cans, as landmarks. \revtext{TurtleBot robot in front of one such landmark is seen in Fig.~\ref{fig:april}}. A ROS-based architecture has been developed to implement the approach. Belief estimation is carried out using EKF. \revtext{We note here that presently we consider static obstacles while planning and therefore the planned trajectories are collision-free. However, to be robust to dynamic obstacles, the plan execution is trivially extended to employ any collision avoidance approach in dynamic environments~\cite{park2018IEEE,zhu2019RAL}. Snapshots of dynamic obstacle avoidance during the execution of a plan can be seen in Fig.~\ref{fig:execution}. As seen in the figure, dynamic obstacles are simulated using TurtleBot robots (white in figure). We now report here the execution time for the scenario discussed in Section~\ref{subsec:scalability}. When 2, 4, 6 number of cubicles are to be visited to collect the corresponding number of documents, the execution times are 140.21$s$ ($\pm$ 3.11$s$), 366.40$s$ ($\pm$ 4.99$s$), and 664.71$s$ ($\pm$ 16.28$s$), respectively. We note here that the execution time varies with robot and its control limits.}

\subsection{Corridor Domain}
\revtext{Our \textit{corridor domain} (see Section~\ref{sec:nav_domain2} for a detailed description) is a variant of the robot navigation domain in~\cite{jiang2019FITEE}. However, they treat it as a task planning problem assuming that feasible motion plans exist for the synthesized task plans. In contrast, we perform task-motion planning}. In this domain of $12m \times 25m$, \revtext{a mobile robot, starting from a given room, navigates an office floor to visit a set of rooms that are selected randomly. The office floor has ten rooms and the robot is initially located in room $1$. All the rooms are connected to each other through the central corridor. In addition, five rooms are directly connected with each other via doors which need to be opened by the robot. The goal is to visit a set of rooms $R$ that are randomly selected for each run}. Since these visits have to be carried out expending as less cost as possible, the robot needs to assess the accessibility between the rooms that are directly connected to each other via a door. This is facilitated through the \texttt{goto\_door} action \revtext{as discussed in Section~\ref{sec:nav_domain2}}. The map of the building floor is as shown in Fig.~\ref{fig:building}.  

\subsubsection{Validation and Scalability}

\begin{figure}[h]
	\centering
		\includegraphics[scale=0.08]{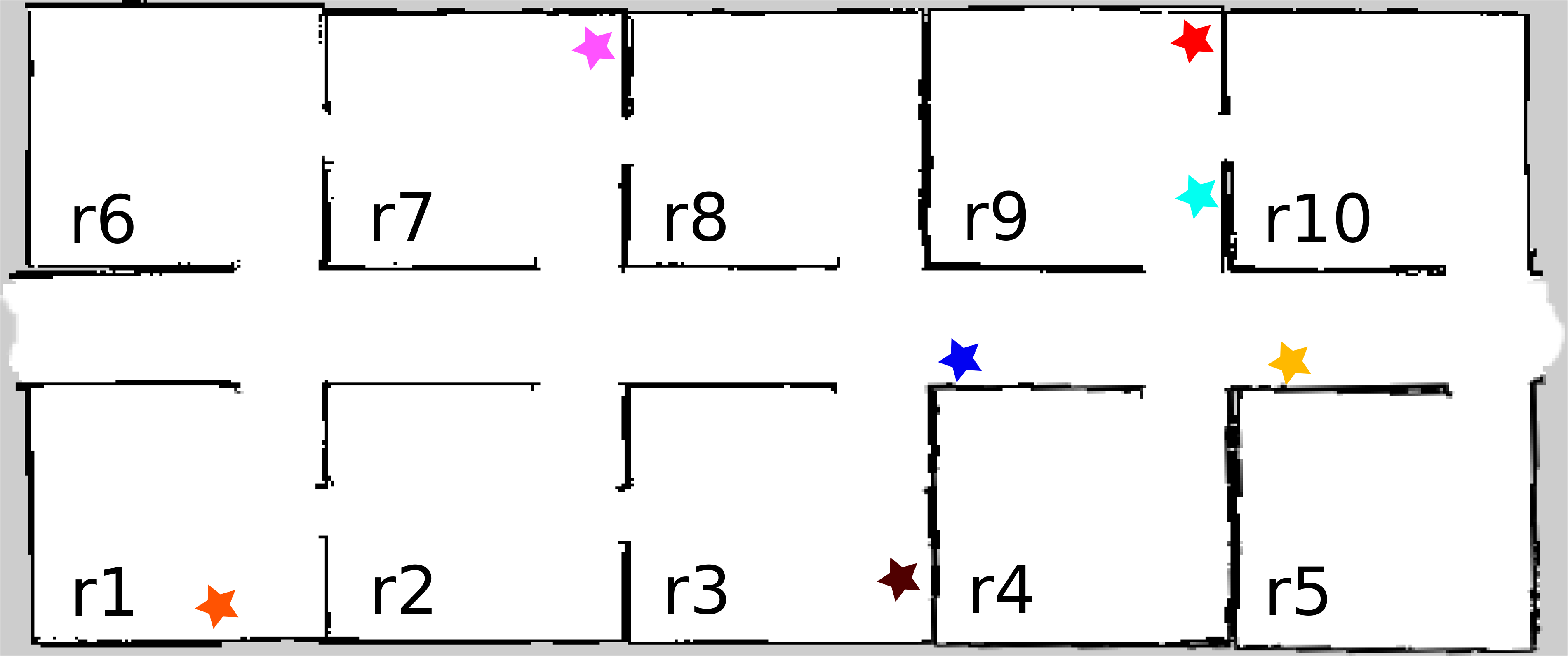}
		\caption{Map of the building floor environment with half the rooms connected directly by doors. The stars with different colors represent landmarks that aids the robot in better localization.}
	\label{fig:building}
\end{figure}

\begin{figure}[h]
	\centering
		\includegraphics[scale=0.5]{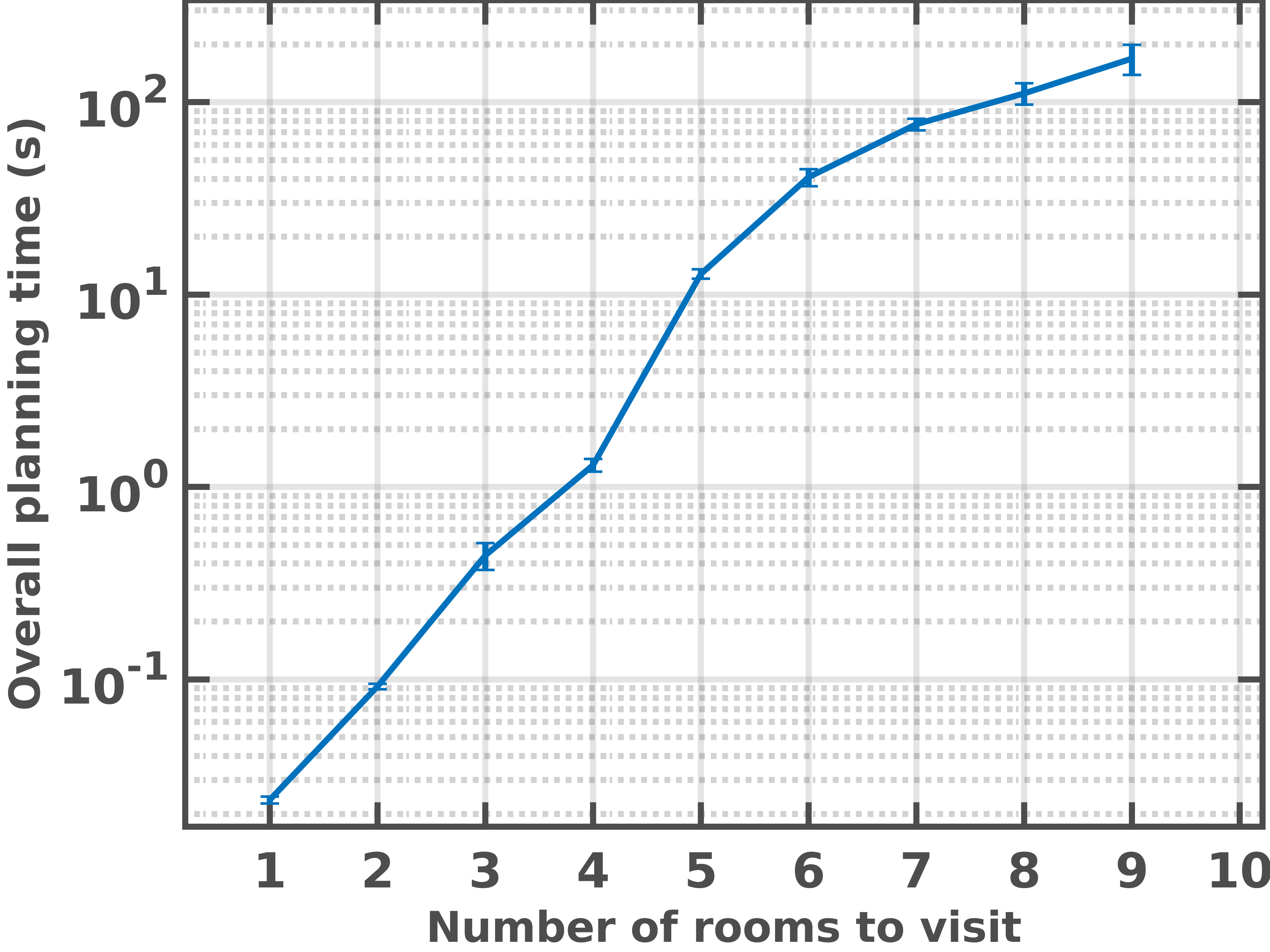}
		\caption{Overall task-motion planning time for different number of rooms that need to be visited in log scale. Planning times are the average for 25 different runs.}
	\label{fig:planning_time}
\end{figure}

\begin{figure}[h]
	\centering
	
\subfloat{\includegraphics[scale=0.33]{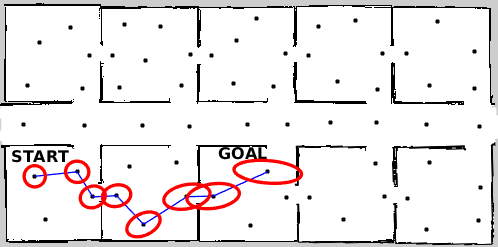}\label{fig:1}}
	\hspace{0.04cm}
\subfloat{\includegraphics[scale=0.33]{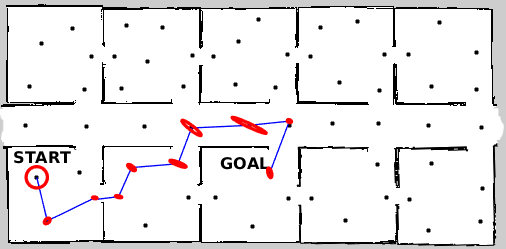}\label{fig:2}}
\vspace{0.05cm}
\subfloat{\includegraphics[scale=0.348]{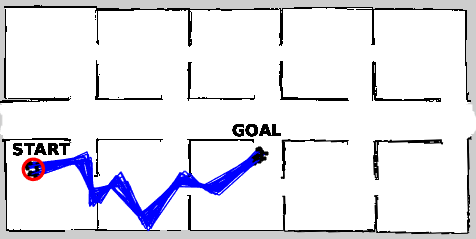}\label{fig:3}}
\hspace{0.04cm}
\subfloat{\includegraphics[scale=0.35]{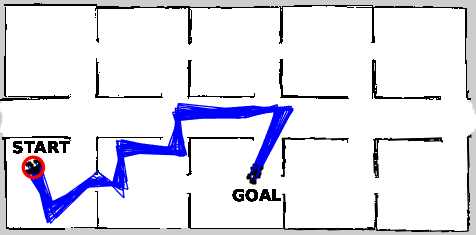}\label{fig:3}}
\vspace{0.05cm}
		\caption{(\textit{top-left} and \textit{top-right}) The propagated belief distributions along the planned paths \revtext{while running MPTP with \textit{PETLON cost} and \textit{MPTP cost}}. The belief estimates for a single planning instantiation corresponding to a unique set of simulated observations are shown. The black dots represent the sampled poses. (\textit{top-left}) Shortest path route that corresponds to \revtext{running the planner with} \revtext{\textit{PETLON cost}}. (\textit{top-right}) Belief space planning corresponding to \revtext{running the planner with} \revtext{\textit{MPTP cost}}, returning an information rich route. (\textit{bottom-left}) Traces of robot's true state while starting from the initial belief and run on \revtext{\textit{PETLON cost}}-- 80\% of the trajectories lead to collision. (\textit{bottom-right}) Traces of robot's true state while starting from the initial belief and run on \revtext{\textit{PETLON cost}}-- only 8\% of the trajectories lead to collision.}
		\label{fig:BSP2}
\end{figure}
First, we run the planner with \revtext{\textit{MPTP cost}}. For a fixed set cardinality $\vert R \vert$ (set elements are the rooms to be visited), 25 trials are performed, where the set elements are selected randomly for each trial. The average planning time for each of them is shown in Fig.~\ref{fig:planning_time}. While the planning time does scale exponentially with $\vert R \vert$, the plan for $\vert R \vert = 9$ is computed in less than 3 minutes. The work in~\cite{jiang2019FITEE} evaluates the task planning performance on a similar domain \revtext{randomly selecting the number of rooms to visit in each trial. Since MPTP performs task-motion planning, the overall MPTP planning times with increasing $\vert R \vert$ is greater than those reported in~\cite{jiang2019FITEE}. However the graph of $\vert R \vert$ with planning time (Fig.~\ref{fig:planning_time}) follows a similar trend to that reported in~\cite{jiang2019FITEE}. It is noteworthy that for a given $\vert R \vert$, the difference in MPTP planning time and the planning time reported in~\cite{jiang2019FITEE} is significantly less.} 

Next, we run the planner with \revtext{\textit{PETLON cost}} and \revtext{\textit{MPTP cost}}. We consider a scenario in which the robot, starting from room $r1$, has to visit rooms $r2$ and $r3$. As seen in Fig.~\ref{fig:building}, rooms $r1$, $r2$ and $r2$, $r3$ are also connected by doors between them. Fig.~\ref{fig:BSP2} on top-left and top-right shows the planned trajectories in both the scenarios with the corresponding covariance estimated at each node (only the ($x$,$y$) portion is shown). Note that the illustrations show a single planning instantiation corresponding to a unique set of simulated observations $Z$. Belief space planning (\revtext{\textit{MPTP cost}}) enables effective localization by returning a route which is rich in sensor information (see Fig.~\ref{fig:BSP2} on top-right). Fig.~\ref{fig:BSP2} on the bottom-left, shows the traces of true robot states for 25 different simulations while running on \revtext{\textit{PETLON cost}}. The initial poses are sampled from the known initial belief distribution. Out of the 25 trials, 20 lead to collision on the walls, giving a success rate of only 20\%. The traces of true robot pose for 25 different simulations while running on \revtext{\textit{MPTP cost}} is shown in Fig.~\ref{fig:BSP2} (bottom-right). Only 2 trials lead to collision, giving a success rate of 92\%.

\begin{figure}[h]
	\centering
		\includegraphics[scale=0.5]{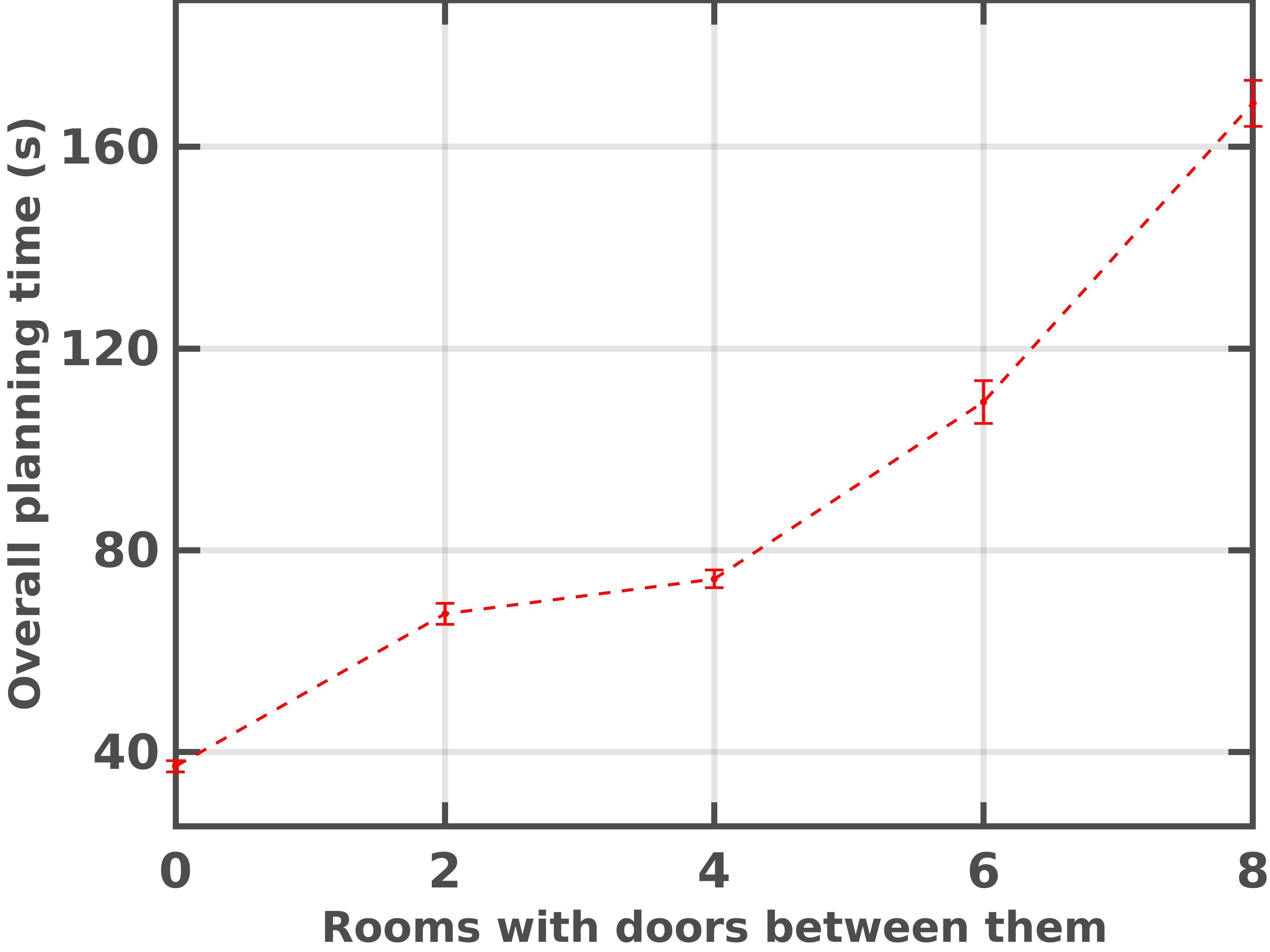}
		\caption{Overall planning time for visiting 7 rooms when the number of rooms directly connected by doors are varying. Average time for 25 trails are plotted in each case.}
	\label{fig:roomDomainScale}
\end{figure}

Finally, we test the scalability of our approach by running the planner with varying number of rooms that are directly connected by doors between them. We consider a scenario in which seven rooms are to be visited. We consider five different cases of this scenario, each of which has a fixed number of rooms that are directly connected by the doors. For each case, 25 trails are performed and for each trial, the rooms with doors between them are randomly selected. The overall planning time is seen in Fig.~\ref{fig:roomDomainScale}. 

%% file: discussion.tex
\revtext{In this section, we first discuss some limitations of our approach and later comment on the relation to multi-goal planning and travelling salesman problems. 

MPTP has few limitations and assumptions and relaxing them would enhance the capability and robustness of our approach in challenging scenarios. First, we sample collision-free poses and therefore considering static obstacles, the planned trajectories are collision-free. In this sense, we employ a deterministic collision avoidance approach and do not compute the probability of collisions while computing a path during planning. It is a reasonable assumption for all practical purposes but is not the case in general while planning in narrow regions or corridors. The execution may be trivially extended to consider collision probabilities, making it robust to both static and dynamic obstacles. Second, we assume straight line path between two sampled poses. This might not fare well in some experimental domains and can lead to larger prediction uncertainties. Presently, as the number of samples vary, the search is performed again. It is our future direction to efficiently utilize the previous search results to reduce the computation time for increased samples. It is also an interesting future direction to extend the framework to an online real-time planning approach.

Multi-Goal Planning (MGP)~\cite{saha2003ICRA}, where a robot visits a sequence of goal configurations is a subset of the general class of TMP problems. Most existing MGP approaches~\cite{saha2003ICRA,imeson2014ICRA,alatartsev2015JIRS,imeson2019TRO} leverage the Traveling Salesman Problem (TSP)~\cite{applegate2006book} solvers for task sequencing. A TSP problem finds a minimum cost path traversing a set of points such that every point is visited once. In an MGP problem these points correspond to the set of goal configurations the robot needs to visit. It can be argued that all MGP problems can be modeled as a TMP problem but not vice versa. For instance, consider the \textit{office domain} presented in Section~\ref{sec:nav_domain1}. In this scenario the robot not only has to visit regions of interest but execute actions such as collecting the documents, which is to be performed when visiting each cubicle ensuring that the action preconditions are met. Moreover, in certain scenarios cubicles may need to be visited multiple times violating the single visit constraint of traditional TSP solvers. The \textit{corridor domain} (see Section~\ref{sec:nav_domain2}) presents additional challenges for TSP solvers. If we consider that there are no doors between the rooms, then the problem reduces to just visiting different rooms and can be solved using TSP solvers. However, in the considered scenario there are doors between certain rooms and the accessibility between the rooms that are directly connected to
each other via a door needs to be assessed by the robot. This requires different levels of reasoning to verify the action preconditions such as, checking if a door exists, navigating to the door, checking if the trace of the robot pose covariance is within the uncertainty budget and if yes, then updating the roadmap. Moreover, if the robot passes through the door, the accomplishment of the action effect (in this case, closing the door corresponds to updating the roadmap) needs to be established. Thus MPTP is able to solve a larger class of problems than traditional TSP solvers.}

%% file: conclusion.tex
This paper introduces an approach for task-motion planning under motion and sensing uncertainty. Task-motion interaction is facilitated by means of semantic attachments that return motion costs to the task planner. In this way, the action costs of the task plans are evaluated using a motion planner. The plan synthesized is optimal at the task-level since the overall action cost is less than that of other task plans generated for a given roadmap. It is to be noted that the action cost also encompasses the motion cost. The proposed approach is probabilistically complete and we have validated the framework using a simulated office environment in Gazebo and a corridor environment. The approach has been evaluated with different configurations that correspond to different motion cost computation, illustrating the need for a combined TMP approach for navigation in belief space. Though we have validated MPTP in two different robot navigation domains, real-world scenarios often require large number of tasks to be performed. Real-world domains are much more knowledge-intensive, significantly increasing the task-level and motion-level complexity. The scalability results suggest that our approach fares well with respect to increased task-level complexity and plan length.